\documentclass[pdflatex,sn-basic]{sn-jnl}

\usepackage{graphicx}%
\usepackage{multirow}%
\usepackage{amsmath,amssymb,amsfonts}%
\usepackage{amsthm}%
\usepackage{mathrsfs}%
\usepackage[title]{appendix}%
\usepackage{xcolor}%
\usepackage{textcomp}%
\usepackage{manyfoot}%
\usepackage{booktabs}%
\usepackage{algorithm}%
\usepackage{algorithmicx}%
\usepackage{algpseudocode}%
\usepackage{listings}%

\usepackage{caption}
\usepackage{subcaption}
\usepackage{url}
\usepackage{booktabs}
\usepackage{makecell}
\usepackage{multirow}
\usepackage{siunitx}

\theoremstyle{thmstyleone}%
\theoremstyle{thmstyletwo}%

\theoremstyle{thmstylethree}%

\begin{document}

\title[Article Title]{Implicit Multi-Camera System Calibration Using Gaussian Processes}


\author*[1]{\fnm{Ivan} \sur{De Boi}}\email{ivan.deboi@uantwerpen.be}
\author[1]{\fnm{Bart} \sur{Ribbens}}\email{bart.ribbens@uantwerpen.be}
\author[]{\fnm{Veronika} \sur{Golanova}}\email{golanova.ve@gmail.com}
\author[]{\fnm{Ursula} \sur{Kapov}}\email{ukapov@gmail.com}
\author[1]{\fnm{Simon} \sur{Verspeek}}\email{simon.verspeek@uantwerpen.be}

\affil[1]{\orgdiv{InViLab}, \orgname{University of Antwerp}, \orgaddress{\street{Groenenborgerlaan 179}, \city{Antwerp}, \postcode{2020}, \country{Belgium}}}

\abstract{This paper proposes a novel framework for implicit multi-camera system calibration utilizing Gaussian Process (GP) regression. Conventional explicit calibration methods are constrained by rigid mathematical models and struggle with complex, non-linear distortions from unconventional optics, while existing neural network-based implicit approaches are typically data-hungry and lack inherent uncertainty quantification (UQ). Our GP-based model directly learns the complex, non-linear mapping from 2D image coordinates across all cameras to a 3D world coordinate, completely bypassing  time-consuming estimation of explicit intrinsic and extrinsic parameters. Moreover, the inherent UQ is critical for transforming a simple 3D point prediction into a verifiable 3D measurement, complete with statistically-sound confidence bounds. To further enhance data efficiency and practical deployment, we integrate Active Learning (AL), which intelligently leverages the GP's predictive uncertainty to strategically guide the acquisition of new calibration data. This approach results in a robust, data-efficient, and reliable calibration solution, proving particularly effective in practical scenarios where collecting extensive calibration data is a dominant constraint. Our experiments show that the uncertainty for the 3D predictions is higher closer to the cameras. The data points in $uv$-coordinate space are more sparse in that region, even though they are not in 3D space. This work is relevant for anyone who is tasked with the calibration of complex multi-camera systems.}




\maketitle

\section{Introduction}
The derivation of accurate 3D world information from multiple cameras is a fundamental objective in computer vision, with applications ranging from people tracking~\citep{Wang2013IntelligentReview, Sakaguchi2024Multi-CameraConsiderations}, object tracking~\citep{Amosa2023Multi-cameraAdvances, Sie2021FieldEstimation}, pose estimation~\citep{Sun2020Multi-viewEstimation, Parisotto2023MORE:Estimation}, Simultaneous Localisation And Mapping (SLAM)~\citep{Yang2020Multi-cameraNavigation, Yang2024MCOV-SLAM:System} and many more.

For multi-camera systems, this objective is critically dependent on the process of camera calibration. The conventional and most widely adopted methodology is explicit calibration, which involves determining a set of parameters that mathematically model the image formation process. These parameters are typically divided into two categories: intrinsic parameters (e.g., focal length, principal point, lens distortion coefficients) that define the internal projective geometry of the camera, and extrinsic parameters that describe the camera's rigid-body transformation (position and orientation) relative to a world coordinate system. Zhang's method~\citep{Zhang2000ACalibration, Burger2016ZhangsImplementation}, which utilises observations of a planar pattern from various viewpoints, remains a ubiquitous and effective method for this task and is implemented in both the MATLAB Calibration Toolbox~\citep{Scaramuzza2006OCamCalib:Matlab, Zhang2020CameraExamples} and the OpenCV library~\citep{Uranishi2018OpenCV:Library}.

Despite its prevalence, explicit calibration presents significant practical challenges, particularly for complex, non-standard imaging systems~\citep{Jiang2023DiscussionSystems, Kobayashi2020WideVision}. Practically, the procedure is laborious, and its ultimate accuracy is acutely dependent on the quality and density of the collected observations. To ensure a precise determination of parameters, a sufficiently dense distribution of feature points within the captured 3D scene is required to prevent degenerate geometric situations that can skew the modelling of the forward imaging process.

A more profound issue arises from the limited flexibility when the imaging deviates strongly from the assumed model (e.g. extreme wide‑angle optics), although they remain highly accurate when their assumptions hold. The conventional pinhole model, for instance, approximates complex optical elements with a single perspective projection, a simplification that suffers from inherent accuracy limitations~\citep{Zhang2020CameraExamples}. This challenge is further exacerbated by unconventional optics, such as wide-angle or fisheye lenses, which introduce severe, high-order, non-linear distortions that simple polynomial models struggle to accurately parametrise. While advanced parametric models have been proposed to account for such complex distortions~\citep{DeBoi2024HowModel, Lochman2021BabelCalib:Cameras}, they still operate within the constraints of fitting a predefined mathematical model, which is a strength in standard, near-pinhole configurations but becomes a weakness when the imaging geometry deviates strongly from those assumptions, such as in highly unconventional, wide-angle multi-camera setups.

To address this constraint while remaining within the explicit framework, research has explored the path of hyper-parametrization. Schöps et al.~\citep{Schops2019WhyTwelve} controversially argued that adopting an unconstrained model with thousands of free parameters, suggesting that having "10,000 parameters" is superior to twelve, yields a higher precision calibration capable of modelling complex, non-perspective imaging geometries. While such highly parametrised models offer increased functional capacity, they are still fundamentally restricted by the necessity of fitting a predefined mathematical structure, inevitably leading to complex and exhaustive non-linear optimisation procedures for parameter determination.

This realization has led to the development of an alternative paradigm known as implicit calibration. This approach circumvents the estimation of physical camera parameters by instead learning a direct functional mapping from a set of 2D image-space coordinates to their corresponding 3D world-space coordinates. Early implementations of this concept utilised artificial neural networks (ANNs) to learn this complex, high-dimensional relationship~\citep{WooDong-Min2006ImplicitNetwork}. However, ANNs typically exhibit significant data requirements and are prone to overfitting. Furthermore, they generally lack a principled mechanism for quantifying the uncertainty associated with their predictions, a notable limitation for applications where reliability is critical.

Uncertainty quantification (UQ) is essential for turning a camera into a validated, high-precision measuring instrument~\citep{Zhu2009UncertaintyCalibration, Reu2013AApproach, Leizea2023CalibrationBudget}. Without UQ, the intrinsic and extrinsic camera parameters are merely point estimates with unknown errors, leading to unreliable results in critical applications. UQ allows for error propagation to the final 3D measurements. UQ also serves as a diagnostic tool, identifying which parts of the calibration setup caused the most noise, which is invaluable for designing more robust systems. This need is amplified in multi-camera systems due to the increased complexity of the extrinsic calibration and the compounding of errors. In these setups, UQ must quantify the uncertainty of each camera's parameters and the critical relative pose (rotation and translation) between every camera pair. Errors in this relative pose directly cause misalignment and scale distortion when fusing multi-view data.

This paper proposes and evaluates a novel framework for implicit multi-camera calibration based on Gaussian Process (GP) regression~\citep{Rasmussen2006}. GPs are a non-parametric, probabilistic machine learning technique that offers a robust alternative to neural network-based approaches. By employing a kernel to measure similarity between data points, GPs can flexibly model complex, non-linear functions without imposing a rigid parametric form. Their Bayesian formulation provides two key advantages for the calibration task: first, a demonstrated capacity for high performance in low-data regimes (less data needed to perform the calibration procedure), and second, the intrinsic ability to provide a statistically meaningful variance for each prediction, thereby quantifying model uncertainty. 

While GPs are effective in low-data scenarios, achieving high-precision calibration across a large or geometrically complex workspace necessitates a strategic approach to data acquisition. The conventional method of manually collecting a comprehensive, dense set of calibration points is both laborious and often results in an inefficient distribution of samples, dedicating resources to regions already well-modelled. To mitigate this inefficiency, we propose the integration of the Active Learning (AL) methodology within our GP framework. AL leverages the predictive variance, which can be interpreted as the quantified uncertainty naturally provided by the GP model, to intelligently direct the data collection process. By prioritizing the sampling of new calibration points in areas exhibiting high model uncertainty, AL ensures that each new observation provides maximum informational gain. This combined GP-AL approach not only yields a robust, probabilistically-grounded implicit calibration model but also substantially reduces the required physical effort and time, thereby establishing a more practical and efficient calibration procedure.

\begin{figure}[!t]
\centering
\includegraphics[width=0.65\textwidth]{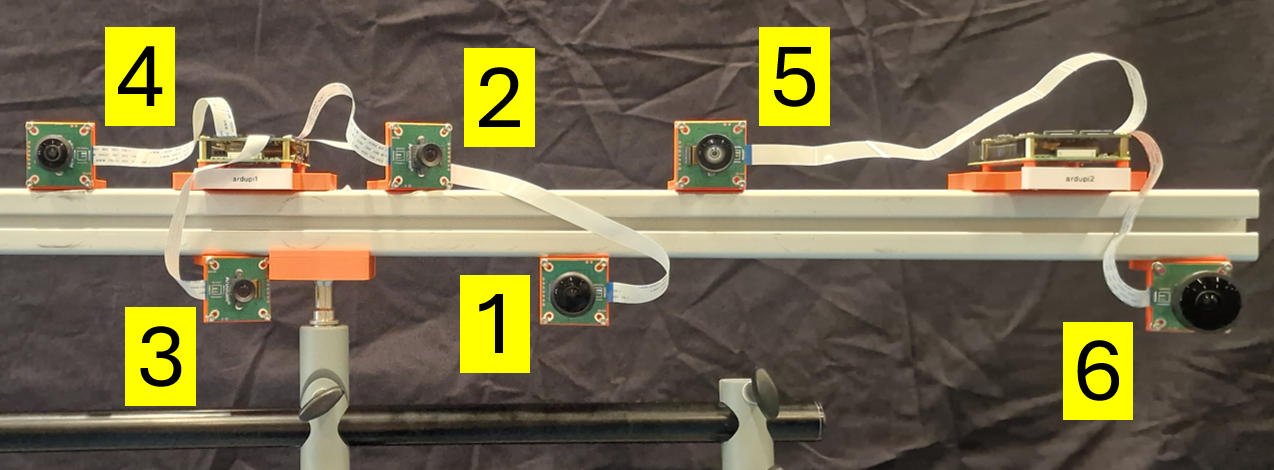}
\caption{Six cameras mounted on a bar. Camera 1, 4, 5 and 6 have various wide-angle lenses. Camera 2 and 3 are regular cameras. Combination of these cameras are used to create three different setups: two regular cameras, two regular plus two wide-angle cameras and two regular plus four wide-angle cameras.}
\label{fig:6Cams}
\end{figure}

\begin{figure}[!t]
\centering
\includegraphics[width=0.65\textwidth]{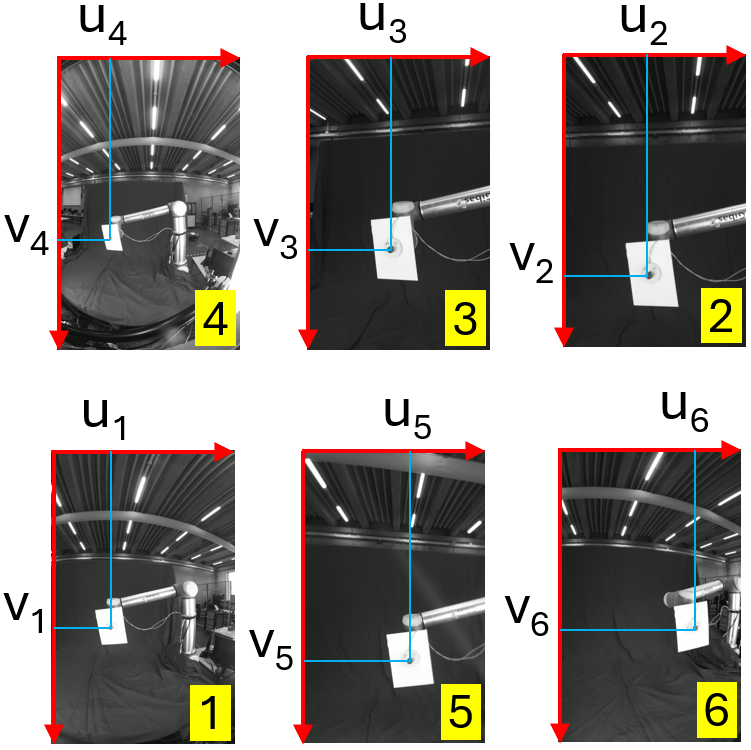}
\caption{Six images produced by the cameras in the setup depicted in Figure \ref{fig:6Cams}. The $uv$-coordinates of an object with known $xyz$-coordinates are used to learn the mapping from the total pixel space to the 3D world.}
\label{fig:6Images}
\end{figure}

In this work, we utilise GPs to directly model the relationship from pixel $uv$-coordinates to corresponding $xyz$-coordinates of points in 3D. Figure \ref{fig:6Cams} shows an example setup of a multi-camera system. These six identical cameras are equipped with lenses with different fields of view (see also Section \ref{sec:Experiments} for more details). The combination of $uv$-coordinates in every image of an object, see Figure \ref{fig:6Images}, and the corresponding known $xyz$-coordinates serve as training data for the GPs. Once the GPs are trained, the relationship between the pixels of the images and the 3D world is learned, and thus the multi-camera system is calibrated. Our proposed method bypasses the explicit calibration of each individual camera, hence the term implicit for the calibration of the multi-camera system as a whole.

Our main contributions are:
     \begin{enumerate}
         \item \textbf{Novel Gaussian Process Based Calibration:} We implement a GP regression model that maps concatenated multi-view pixel coordinates to 3D locations, sidestepping explicit intrinsic and extrinsic estimation.
         \item \textbf{Uncertainty Quantification of the Calibration:} We expose the GP predictive variance as a calibration quality measure and visualise it in 3D.
         \item \textbf{Active Learning for Efficiency:} We couple GP variance with an uncertainty sampling scheme to select new calibration views.
     \end{enumerate}


\section{Related work}\label{sec:RelatedWork}

Contemporary research is increasingly advancing non-parametric and implicit representations, often integrating deep learning architectures to tackle multi-view geometry problems. Modern non-parametric models aim to relate each pixel directly to its corresponding 3D observation ray, thereby overcoming the inherent limitations of standard parametric forms in complex multi-camera setups~\citep{Amosa2023Multi-cameraAdvances}. Deep learning methods have introduced novel calibration targets, such as pixel-wise geometry fields (e.g., camera rays or distortion distribution maps), effectively replacing the rigid traditional intrinsic and extrinsic parameters with a learning-friendly, pixel-level parametrization~\citep{Badrinarayanan2017SegNet:Segmentation, Peng2022PVNet:Estimation}. This concept has been powerfully extended by integrating implicit representations with volumetric reconstruction. Specifically, architectures like Self-calibration NeRF~\citep{Jeong2021Self-CalibratingFields} allow for the simultaneous self-calibration of generic cameras—even those exhibiting arbitrary non-linear distortions—during the process of 3D structure reconstruction and novel view synthesis across multiple views, fitting all camera types through pixel-level regression.

Our proposed method relies on Gaussian processes, a non-parametric, probabilistic model that defines a distribution over functions rather than just parameters. It uses a kernel function (covariance function) to learn complex, non-linear relationships directly from the data. Section \ref{ssec:GPs} provides a more detailed explanation. This allows it to represent model discrepancy (the error due to an imperfect camera model) and the effects of complex noise much more flexibly than deep neural networks.  GPs inherently provide a predictive distribution (a mean and variance) for any unobserved output (e.g., a new 3D point or a new image coordinate) given its input. This uncertainty estimate is often more calibrated and directly incorporates the uncertainty from both the noise (aleatoric uncertainty) and the lack of data coverage (epistemic uncertainty). In essence, it tells you: "How certain am I about the measurement itself?"

In this work, we distinguish between three primary calibration methodologies: explicit, implicit with a neural network, and implicit with a GP. The conventional explicit paradigm aims to determine specific physical parameters, such as the intrinsic and extrinsic parameters. This approach relies on rigid, predefined mathematical constructs, typically the pinhole model augmented with polynomial distortion correction. While mathematically grounded, its primary weakness lies in its inherent sensitivity to highly complex, non-linear distortion introduced by unconventional optics like wide-angle lenses. Furthermore, its ability to quantify uncertainty is non-intrinsic, generally relying on a post-hoc analysis of residual errors~\citep{Lin2007ComparingCalibrations}. 

In contrast, the implicit with a neural network approach shifts the objective entirely, learning a direct, highly flexible, and deterministic function to map 2D image coordinates to 3D world coordinates. This non-parametric flexibility is achieved through powerful function approximators, but it introduces major systemic drawbacks: the method is notably data-hungry, prone to overfitting the complex 2D-to-3D relationship, and critically, lacks a principled mechanism for uncertainty quantification. Even when implemented, UQ for neural networks is often ad-hoc (e.g., using techniques like dropout or bootstrapping) and still suffers from generating overconfident point estimates, particularly when extrapolating beyond training data~\citep{Guo2024ALearning}.  

Finally, the implicit with a Gaussian process paradigm retains the non-parametric flexibility of implicit modelling but elevates it to a Bayesian framework. Its core objective is to learn a direct probabilistic function for the 2D to 3D mapping. Characterised by a non-parametric function distribution, its main practical limitation is that computational complexity can scale poorly with extremely large datasets. This is not a problem in our context, as we aim to calibrate our multi-camera setups with as few data points as possible. Crucially, its Bayesian foundation provides intrinsic uncertainty quantification, yielding statistically meaningful variance that cleanly separates epistemic and aleatoric sources. This reliability transforms the calibration output from a simple point prediction into a verifiable 3D measurement complete with confidence bounds, making the GP framework uniquely suited for safety-critical applications.

GPs have been applied to camera calibration across three primary scenarios, all falling under the explicit paradigm with added UQ. First, Hillen et al.~\citep{Hillen2023EnhancedProcesses} leveraged GPs to reconstruct and enhance the calibration target. They learned a mapping from a known grid of virtual corners on a checkerboard to the observed pixel coordinates, which allowed them to infer missing corners and perform sub-pixel enhancement of corner locations. Second, GPs have been directly used to model the highly non-linear lens distortion function itself~\citep{Ranganathan2012GaussianModeling}, providing a flexible, non-parametric alternative to traditional polynomial models. Third, in~\citep{DeBoi2024HowModel}, a GP was implemented as a preprocessing step to effectively "undistort" any camera, transforming it into a mathematically perfect pinhole model before the main estimation process. To the best of our knowledge, while these methods explicitly address components of camera calibration, our work is the first to bypass the explicit camera calibration of a multi-camera system with a Gaussian process. Furthermore, we are unaware of any work being done on camera calibration based on AL.

\section{Methods}\label{sec:Methods}

\subsection{Gaussian Processes}\label{ssec:GPs}

Gaussian Processes (GPs) establish a robust probabilistic framework for modelling non-linear relationships. A comprehensive exploration of this topic is available in the seminal text by Rasmussen and Williams~\citep{Rasmussen2006}. We construct a training set defined by the pair ${\mathbf{X},\boldsymbol{\mathbf{y}}}$ from n samples, where
$\mathbf{X} = \begin{bmatrix}
	\boldsymbol{\mathbf{x}}_{1}, \boldsymbol{\mathbf{x}}_{2},..., \boldsymbol{\mathbf{x}}_{n}
\end{bmatrix}^T$
is an $n\times d$ matrix of $n$ input vectors of dimension $d$ and
$\boldsymbol{\mathbf{y}} =\begin{bmatrix}
	y_{1}, y_{2},..., y_{n}
\end{bmatrix}^T$
is a vector of continuous-valued scalar outputs. These are the training points. Our objective in regression is to discover a function $f:\mathbb{R}^d\rightarrow\mathbb{R}$,
\begin{equation}\label{GPR}
	y = f(\boldsymbol{\mathbf{x}})+\epsilon, \quad \epsilon\sim\mathcal{N}(0,\sigma_{\epsilon}^2) ,
\end{equation}
with $\epsilon$ being Gaussian noise. We model this function f using a Gaussian process, which is completely specified by its mean function $m(x)$ and its covariance function $k(\boldsymbol{\mathbf{x}},\boldsymbol{\mathbf{x}}')$. This covariance function, often called the kernel, is parametrised by a vector of characteristic values known as hyperparameters, $\theta$.

By its fundamental definition, a GP places a probability distribution over a collection of functions, meaning that any finite set of function values will be jointly normally distributed:
\begin{equation}\label{joint f f*} 	 	
	\begin{bmatrix}
		\mathbf{f}\\
		\mathbf{f_{*}}
	\end{bmatrix}
	\sim		
	\mathcal{N}		
	\left(
	\begin{bmatrix}
		\boldsymbol{m}_{\mathbf{X}}\\
		\boldsymbol{m}_{\mathbf{X}_{*}}
	\end{bmatrix}
	,
	\begin{bmatrix}
		\mathbf{K}_{\mathbf{X},\mathbf{X}} & \mathbf{K}_{\mathbf{X},\mathbf{X}_{*}}\\
		\mathbf{K}_{\mathbf{X}_{*},\mathbf{X}} & \mathbf{K}_{\mathbf{X}_{*},\mathbf{X}_{*}} 
	\end{bmatrix} 
	\right) 		,
\end{equation}
where $\mathbf{X}$ and $\mathbf{X}_{*}$ are the input vectors of the $n$ observed training points and the $n_{*}$ unobserved test points, respectively. The vectors of mean values for $\mathbf{X}$ and $\mathbf{X}_{*}$ are given by $\boldsymbol{m}_{\mathbf{X}}$ and $\boldsymbol{m}_{\mathbf{X}_{*}}$.
The covariance matrices $\mathbf{K}_{\mathbf{X},\mathbf{X}}$, $\mathbf{K}_{\mathbf{X}_{*},\mathbf{X}_{*}}$, $\mathbf{K}_{\mathbf{X}_{*},\mathbf{X}}$ and $\mathbf{K}_{\mathbf{X},\mathbf{X}_{*}}$ are constructed by evaluating $k$ at their respective pairs of points.
In practice, we do not have access to the latent function values directly, which are dependent on the noisy observations $\boldsymbol{\mathbf{y}}$.

The posterior distribution, conditioned on the data, for the unobserved test function values $\mathbf{f_{*}}$ is formulated as:
\begin{equation}\label{pred f f* final}	 	
	\mathbf{f_{*}}|\mathbf{X}, \mathbf{X}_{*}, \boldsymbol{\mathbf{y}}, \boldsymbol{\theta}, \sigma_{\epsilon}^2
	\sim		 
	\mathcal{N}\left( \mathbb{E}(\mathbf{f_{*}}), \mathbb{V}(\mathbf{f_{*}}) \right)
\end{equation}
\begin{equation}\label{E}	 	
	\mathbb{E}(\mathbf{f_{*}}) = \boldsymbol{m}_{\mathbf{X}_{*}} + \mathbf{K}_{\mathbf{X}_{*},\mathbf{X}}\left[ \mathbf{K}_{\mathbf{X},\mathbf{X}}+\sigma_{\epsilon}^2I\right]^{-1}\mathbf{y}
\end{equation}
\begin{equation}\label{V}	 	
	\mathbb{V}(\mathbf{f_{*}}) = \mathbf{K}_{\mathbf{X}_{*},\mathbf{X}_{*}}-\mathbf{K}_{\mathbf{X}_{*},\mathbf{X}}\left[ \mathbf{K}_{\mathbf{X},\mathbf{X}}+\sigma_{\epsilon}^2I\right] ^{-1}\mathbf{K}_{\mathbf{X},\mathbf{X}_{*}} .
\end{equation}

The choice of kernel is crucial as it dictates the smoothness and adaptability of the GP model. The squared exponential kernel is frequently selected because it is infinitely differentiable, guaranteeing very smooth function outputs. It is defined by the equation:
\begin{equation}	 	
	k_{\mathrm{SE}}(\mathbf{x}, \mathbf{x}') = \sigma^2_{f}\exp \left( -\frac{\left|\mathbf{x}-\mathbf{x}' \right|^2 }{2l^2}\right) ,
    \label{eq:SE} 
\end{equation}
where $\sigma^2_{f}$ is the outputscale and $l$ is the lengthscale that determines the radius of influence of the training points. For the squared exponential kernel the hyperparameters are $\sigma^2_{f}$ and $l$. These are learned in a Bayesian way from the data by maximising the marginal likelihood. For more details, we refer to the book by Rasmussen and Williams~\citep{Rasmussen2006}.

Alternatively, one can assign a distinct lengthscale parameter to each input dimension, a technique termed Automatic Relevance Determination (ARD). This allows the function's correlation structure to vary across different input axes. We can modify Equation~\ref{eq:SE} for the squared exponential kernel to incorporate ARD:
\begin{equation}	 	
	k_{\mathrm{SEARD}}(\mathbf{x}, \mathbf{x}') = \sigma^2_{f}\exp \left(- \frac{1}{2} \sum_{j=1}^d \left( \frac{\left|\mathbf{x}_{j} - \mathbf{x}'_{j}\right|}{l_j} \right)^2 \right) ,
	\label{eq:SEARD}
\end{equation}
in which 
$l_{j}$ is a separate lengthscale parameter for each of the $d$ input dimensions.

Training a GP involves finding the values for the kernel hyperparameters $\theta$ that maximise the log marginal likelihood (also known as the evidence). For Gaussian distributions, the expression for this optimisation target is known in closed form~\citep{Rasmussen2006}:
\begin{equation}\label{eq:mll}	 	
	\log p(\boldsymbol{\mathbf{y}}|\boldsymbol{\theta},\mathbf{X})\propto
    -\dfrac{1}{2} \mathbf{y}^T\left[ \mathbf{K}_{\mathbf{X},\mathbf{X}}+\sigma_{\epsilon}^2I\right]^{-1}\mathbf{y}
 -\dfrac{1}{2} \log |\mathbf{K}_{\mathbf{X},\mathbf{X}}+\sigma_{\epsilon}^2I| .
\end{equation}
We refer to Equation~\ref{eq:mll} as the log marginal likelihood because it is obtained by integrating out the latent function. The principle behind maximizing this value is related to Occam's razor: it balances model fit against complexity. The first term is a quadratic measure of how well the model fits the data y. The second term acts as a penalty on complexity. This term discourages models that fit the data too perfectly but generalise poorly, or models that rely on a measurement noise $\epsilon$ with large variance $\sigma_{\epsilon}^2$.

\subsection{Active Learning}\label{ssec:ActiveLearning}

The Gaussian process posterior variance has several benefits. Besides the fact that this can be interpreted as an uncertainty quantification, it can also be exploited in Active Learning (AL)\cite{Pasolli2011GaussianScheme, Cacciarelli2024RobustLearning}. This is an iterative strategy designed to learn the entire underlying function efficiently by strategically choosing which data points to measure next. This has been studied in various application domains where the underlying function of study is expensive to evaluate, such as parameter design in thermography~\citep{Verspeek2022DynamicEmulation}, diagnosing patients in rehabilitation sciences~\citep{DeBoi2024AssessmentRegression} or the study of rainfall~\citep{Workneh2024ComparisonEthiopia}. 

It leverages the GP's capacity for probabilistic prediction to guide data acquisition. At each step, the trained GP provides a predictive mean, which is the model's best guess for the function value, and a predictive variance, which quantifies the model's epistemic uncertainty across the input space. The core principle is that new data should be acquired where it will be most informative. To learn the whole function, the primary strategy is Uncertainty Sampling, often implemented by selecting a candidate point that maximises the predictive variance. In other words, where the uncertainty is highest. This prioritises exploration by directing resources to areas of the function space where the model's confidence is lowest due to a scarcity of neighbouring training data. Once the point with maximum uncertainty is identified, the system queries an external oracle (the source of truth) for the actual output. This new, labelled data pair is then added to the training set, and the GP is retrained, causing the predictive variance to collapse around the newly acquired point and propagate through the kernel function to neighbouring areas, thus systematically improving the model's fidelity across the entire domain with minimal effort. This process repeats until the overall uncertainty falls below a desired threshold, converges or a predefined data budget is exhausted. 

Various alternatives to Uncertainty Sampling exist, such as Integrated Variance Reduction~\citep{Avramidis1996IntegratedSimulation}, but are outside the scope of this paper.

In this work, we aim to perform the calibration of a multi-camera system as fast as possible, meaning with as few data points as possible. The AL framework based on GP regression lends itself perfectly for this, by exploiting the posterior variance.

\subsection{Explicit Camera Calibration}\label{ssec:Expl}

Explicit camera calibration is the traditional, mathematically rigorous paradigm for
defining the geometric relationship between a camera's 2D image plane and the 3D
world space. It directly estimates a fixed set of physical parameters, divided into
intrinsic and extrinsic groups. The intrinsic parameters characterise the internal
optics, including focal length, principal point coordinates, and coefficients for lens
distortion (radial and tangential). These are typically collected in a $3\times3$
camera matrix $\mathbf{K}$, together with a distortion parameter vector $\mathbf{d}$.
In the standard pinhole model, $\mathbf{K}$ maps ideal, undistorted normalised image
coordinates to pixel coordinates, while lens distortion is modelled as an additional
mapping, parametrised by $\mathbf{d}$, that warps these normalised coordinates
before projection. The extrinsic parameters define the camera's pose in the global
reference system via a rotation matrix $\mathbf{R}$ and a translation vector
$\mathbf{t}$. The entire process typically relies on observing a known calibration
target (e.g., a planar pattern) from multiple views, with the most widely adopted
practical approach being Zhang's method, which efficiently computes these parameters
by minimising the reprojection error using non-linear optimisation techniques
\citep{Zhang2000ACalibration, Zhang2020CameraExamples, Burger2016ZhangsImplementation}.

Stereo calibration builds directly upon the explicit calibration of individual cameras to define the precise geometric relationship between two or more camera coordinate systems, which is the prerequisite for accurate 3D reconstruction. After determining the intrinsic parameters for each camera separately the primary goal shifts to finding the relative pose between them. This extrinsic relationship is characterised by a $3\times3$ rotation matrix $\mathbf{R}$ and a $3\times1$ translation vector $\mathbf{t}$ that transforms coordinates from the first camera's frame to the second's. The final stereo process involves a global non-linear optimisation where a known calibration target is simultaneously observed across multiple poses by both cameras. This joint minimization of the total reprojection error simultaneously solves for $\mathbf{R}$ and $\mathbf{t}$, ensuring a geometrically coupled set of parameters essential for reliable triangulation.

Multi-camera systems rely on simultaneous bundle adjustment~\citep{Triggs2000BundleSynthesis}. This is a global, non-linear optimisation routine that minimises the total reprojection error across all points observed by all N cameras concurrently. The result is a system that is geometrically closed and internally consistent, where the pose of any camera relative to any other camera is derived from their shared relationship with the stable world frame. This joint optimisation is essential for transforming a collection of individual cameras into a single, high-precision instrument capable of large-scale, unified 3D reconstruction. Again we refer to the book~\citep{Hartley2004MultipleVision} for a full explanation, the details of which are outside the scope of this paper.

\subsection{Implicit Camera Calibration}\label{ssec:Impl}

Implicit camera calibration bypasses the explicit camera calibration by learning the relationship from pixel $uv$-coordinates to corresponding $xyz$-coordinates of points in 3D directly. In our proposed method, the implicit with a GP paradigm, we implement 3 GPs in parallel: one for the $x$-component of the 3D location vector, one for $y$-component and one for the $z$-component. These components are independent of each other and so are the GPs. As every camera produces images with 2D $uv$-coordinates, we have 2 $\times$ the number of cameras as input dimension $d$ for each of the three GPs. For $i$ cameras, we write:

\begin{equation}\label{GPxyz}
  \begin{gathered}
    GP_x: \left\{u_1, v_1, u_2, v_2, ..., u_i, v_i\right\} \rightarrow x \\
    GP_y: \left\{u_1, v_1, u_2, v_2, ..., u_i, v_i\right\} \rightarrow y \\
    GP_z: \left\{u_1, v_1, u_2, v_2, ..., u_i, v_i\right\} \rightarrow z .
  \end{gathered}
\end{equation}

Alternatively, this pixels-to-3D relationship can also be captured by a Multi-Layer Perceptron (MLP)~\citep{Guo2024ALearning}. This feedforward neural network consists of an input layer, one or more hidden layers, and an output layer. It processes information strictly forward, with each hidden neuron applying a weighted sum and a non-linear activation function to model complex relationships. The network is trained using backpropagation to adjust its weights and minimise prediction error. This approach is referred to as implicit with a neural network in this text.

\section{Experiments}\label{sec:Experiments}

To assess our proposed implicit method with a GP approach, we constructed a setup with six cameras, as depicted in Figure \ref{fig:6Cams}. These cameras are mounted on a horizontal bar and are kept fixed with respect to each other. Only two of them have a regular lens, while four of them are equipped with a wide-angle lens. These cameras are synchronised for simultaneous image capture via trigger signals received over a local network by two connected Arduino systems (Arduino, Turin, Italy). The six cameras are Arducam 5MP cameras (Arducam, Shenzhen, China), which we setup to take $800\times1280$ pixel images. The six lenses have an effective focal length in millimetre of 1.55, 2.8, 2.8, 1.85, 2.2 and 1.3 respectively. 

For the explicit methods described above, we rely on the implementation of the library OpenCV~\citep{Uranishi2018OpenCV:Library}, version 4.11.0. To provide a robust parametric baseline, cameras with standard focal lengths were calibrated using the Brown-Conrady model, while wide-angle and fisheye lenses were calibrated using the Kannala-Brandt model via OpenCV’s fisheye interface to account for high-degree non-linear distortion. The corners are detected with PyCBD~\citep{Hillen2023EnhancedProcesses}. The GPs are implemented in GPyTorch ~\citep{Gardner2018Gpytorch:Acceleration}. 

To calibrate the six cameras individually, we take 50 images with each of a checkerboard and then perform Zhang's method with OpenCV. Next we take 10 images of the checkerboard for every pair of cameras, with the checkerboard fully visible in both. These images allow us to calibrate the stereo or multi-camera system explicitly. We repeated this process twice: once with 25 images per camera and 6 per pair and once with 10 images per camera and 4 per pair. These fewer images are a random subset of the original 50 images per camera and the 10 images per pair. We performed the explicit camera calibration on subsets of images 10 times and took the median of the overall reprojection errors of the corners. This gives us three calibrated setups: 50 plus 10, median of 25 plus 6 and median of 10 plus 4.

The calibration target, shown in Figure ~\ref{fig:MovCB}, was printed on a Dibond substrate. This Aluminum Composite Material (ACM) consists of a solid polyethylene core sandwiched between two thin aluminum sheets. This construction was selected for its high flexural rigidity and resistance to warping, ensuring the geometric integrity of the checkerboard pattern remains consistent throughout the experimental procedure.

\subsection{Moving Ball in Grid}\label{ssec:MovingBallGrid}

\begin{figure}[!t]
\centering
\includegraphics[width=0.65\textwidth]{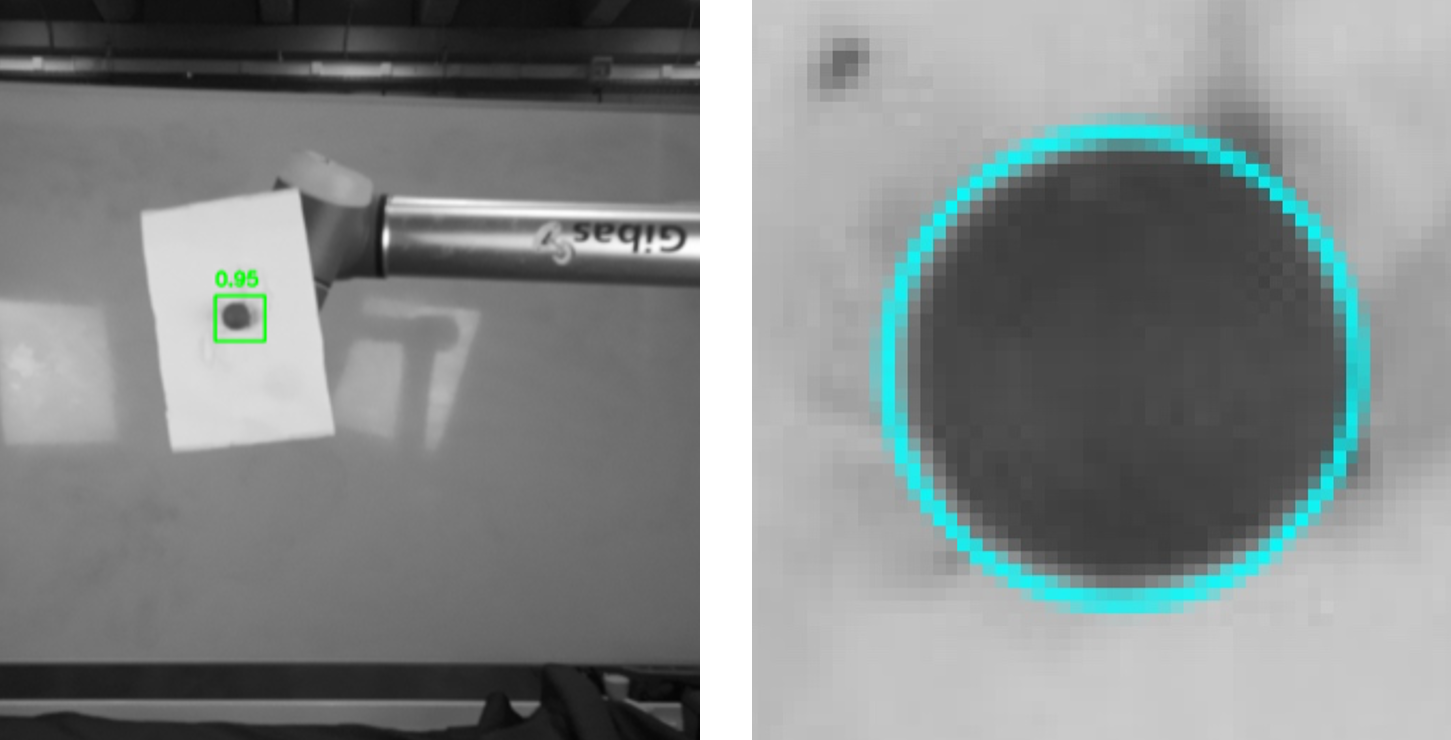}
\caption{Left: A ball mounted on a UR10 robot. Right: a blob detected in the image, after template matching, the region of interest is reduced to the white rectangular area.}
\label{fig:Blob}
\end{figure}

\begin{figure}[!t]
\centering
\includegraphics[width=0.65\textwidth]{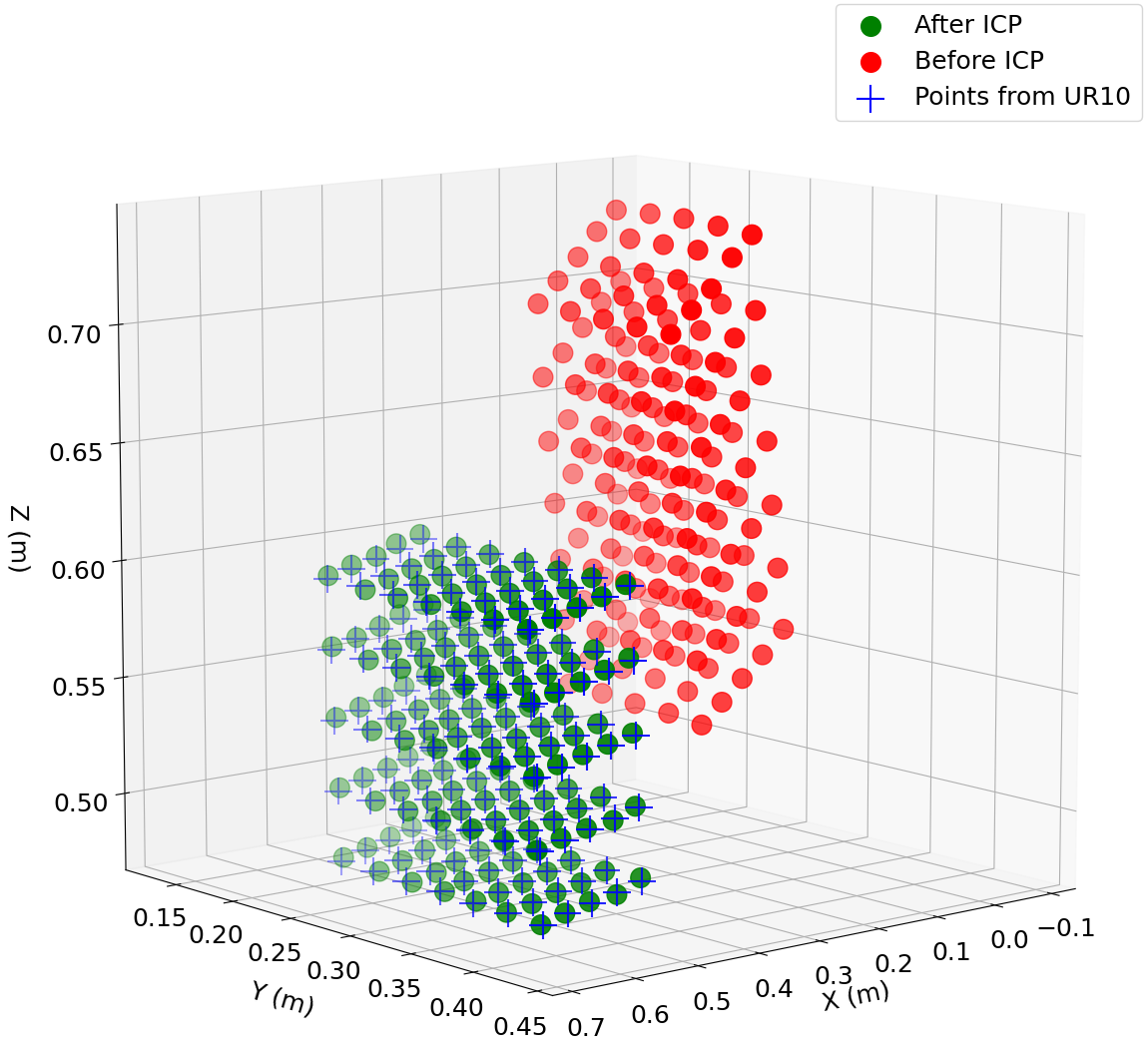}
\caption{The $5\times5\times7=175$ 3D locations of the ball calculated by the explicit model for the dataset with two regular cameras. These are in the reference system of the camera with number 2. We map them to the robot reference system by the iterative closest point algorithm. The blue crosses are the positions the robot recorded and serve as ground truth. Afterwards, the root mean square error between this ground truth and the aligned points in green is calculated.}
\label{fig:ICP}
\end{figure}

 We place the camera bar in front of a working volume of a UR10 robot (Universal Robots, Odense, Denmark). At its tool centre point, we mount a small ball, as shown in Figure \ref{fig:Blob} on the left. The robot moves the ball around while it records the $xyz$-coordinates of the 3D positions. These positions will serve as our ground truth. We place the ball in a regular $5\times5\times7$ grid of 175 3D locations and each time take an image with every camera. The grid is visualised in Figure \ref{fig:ICP} by the blue crosses.

Detection of the ball's $uv$-coordinates is performed in two steps to minimise false positives of blobs in the background. First, OpenCV's template matching is used to crop the image to a smaller region of interest where the ball is expected. The ball is mounted in front of a white rectangular plane, which is what the template is based upon. Second, blob detection is executed only within this reduced region, as shown in Figure \ref{fig:Blob} on the right. The centre of the blob is the $uv$-coordinate of the ball detected in the image. This procedure results in a $uv$-coordinate for every image of the ball taken with every camera. We compose a dataset by corresponding the $uv$-coordinates for $i$ cameras $\left\{u_1, v_1, u_2, v_2, ..., u_i, v_i\right\}$ to their respective $(x, y, z)$ 3-tuples.

We do this for three different sets of cameras taken from the six cameras as depicted in Figure \ref{fig:6Cams}: the two regular cameras (number 2 and 3), two regular and two wide angle cameras (numbers 2, 3, 4 and 5), and the combination of all six cameras (numbers 1 through 6). We denote these datasets as 2R, 2R2W and 2R4W respectively. These dataset have either 2, 4 or 6 cameras, and include the ordinary stereo setup (2R) as well as a mixture of regular and wide-angle lens cameras (2R2W and 2R4W).

For the implicit with a neural network method, we implement a Multi-Layer Perceptron (MLP). As we are dealing with a relatively low number of data points, we take precautions against overfitting by including dropout. The overall structure is a fully-connected deep network that begins with an input layer of size $2\times$ the number of cameras. This input is passed to several fully connected layers with hidden dimension $128\rightarrow128\rightarrow64\rightarrow3$. Between each layer, we apply a leaky RELU and a 0.2 dropout.

To validate the methods, we split these three datasets at random into training and testing parts. The models are validated on the ground truth data in the test set. We apply 90/10, 80/20, 70/30, 60/40, 50/50, 40/60, 30/70, 20/80 and 10/90 for the train-test ratios. This ranges from a dense grid of 157 known 3D locations to a sparse grid of only 17. The explicit method uses the calibrated cameras and the camera system as a whole to calculate the 3D coordinates of the ball based on the detected $uv$-coordinates. It does not use the training data, as there is no training involved. The implicit calibration with a GP and the implicit calibration with a neural network method both rely on the training data to train their underlying model. This means that the GP and the neural network results are already in the reference system of the $xyz$-coordinates provided by the robot. The explicit model provides calculated $xyz$-coordinates in the reference system of one of the cameras. In order to map the two point clouds to the same reference system, we apply the iterative closest point algorithm~\citep{Zhang2022FastPoint}. Figure \ref{fig:ICP} shows a situational overview. 

\begin{figure} 
    \centering
  \subfloat[RMSEs dataset 2R\label{fig:RMSEsa}]{%
       \includegraphics[width=0.65\textwidth]{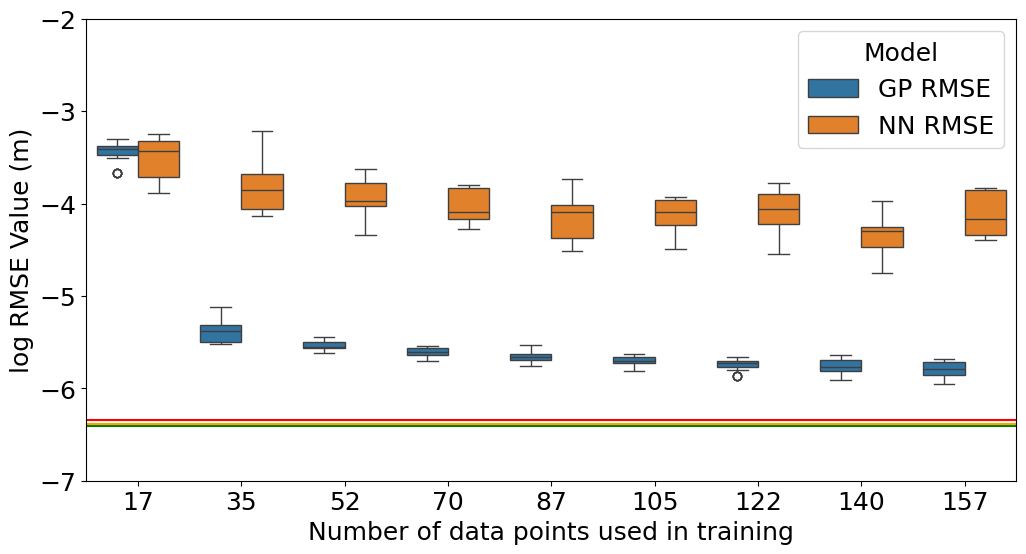}}
    \\
  \subfloat[RMSEs dataset 2R2W\label{fig:RMSEsb}]{%
        \includegraphics[width=0.65\textwidth]{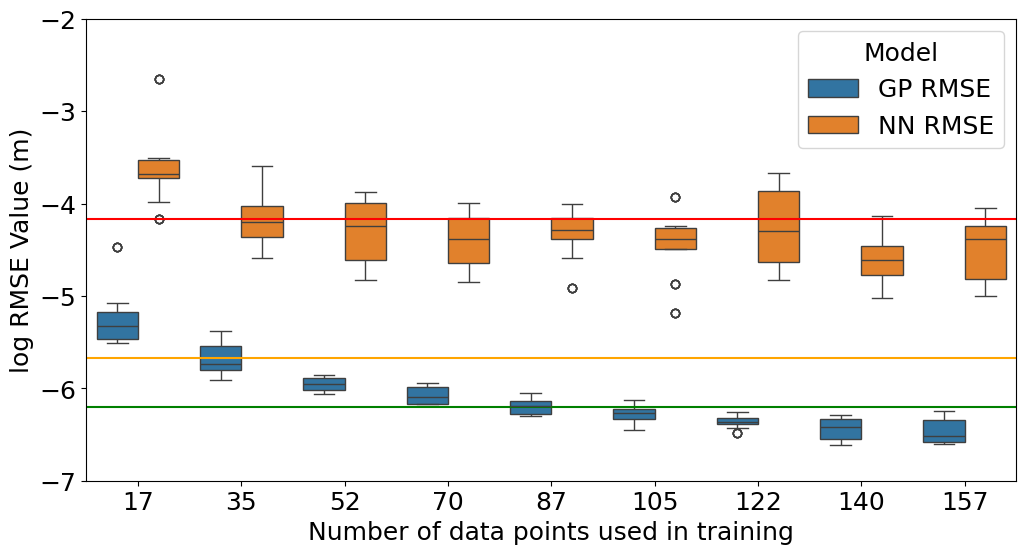}}
    \\
    \subfloat[RMSEs dataset 2R4W\label{fig:RMSEsc}]{%
        \includegraphics[width=0.65\textwidth]{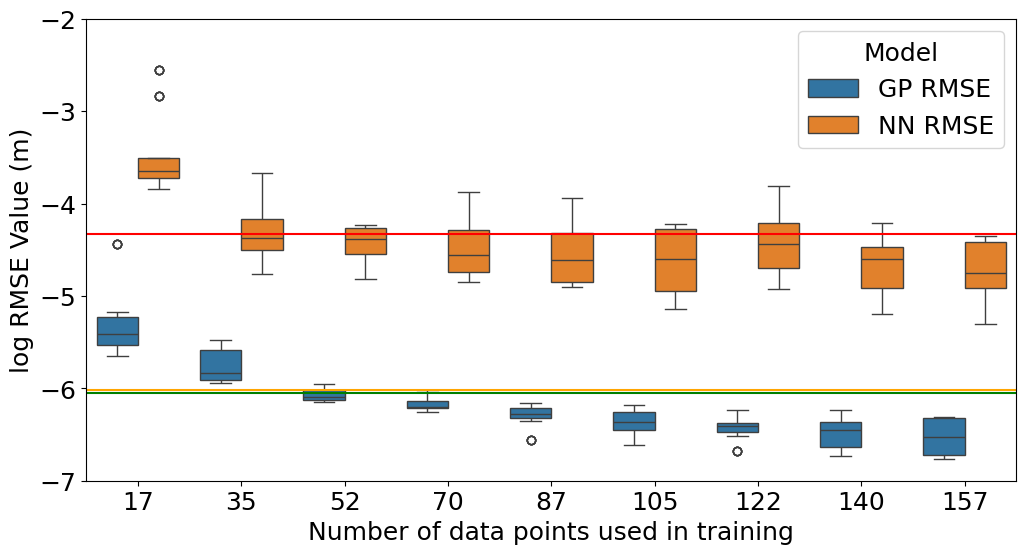}}
    \\
  \caption{The RMSE for the Gaussian process and the neural network model, averaged over all test points, for each of the 50 runs for the three datasets. The green lines are the RMSE of the explicit camera calibration with all 50 images per camera and 10 images per pair. The orange lines are for 25 images per camera and 6 images per pair and the red for 10 images per camera and 4 images per pair.}
  \label{fig:RMSEs} 
\end{figure}

We perform 50 runs for each unique combination of dataset and train-test split ratio, resulting in 1350 runs in total. For every run, we calculate a root mean square error (RMSE) between the ground truth 3D locations provided by the robot and the the calculated (explicit) and predicted (implicit) $xyz$-coordinates for the validation data. The explicit method does not rely on these datasets for training, so we calculate the average RMSE for all 175 3D locations for every dataset, which serves as a baseline to compare against. Afterwards, the average RMSE for all test points is calculated, yielding one RMSE value for the GP and the neural network for one run. The results for all runs and all train-test ratios (in number of data points) are plotted in Figure \ref{fig:RMSEs}. We take the log of the RMSE values for reasons of clarity in the plots, as we wish to simultaneously visualise the higher neural network RMSE values and still keep a level of detail for the lowest values of the GP in the regions with more data points.

\begin{figure}[!t]
\centering
\includegraphics[width=0.65\textwidth]{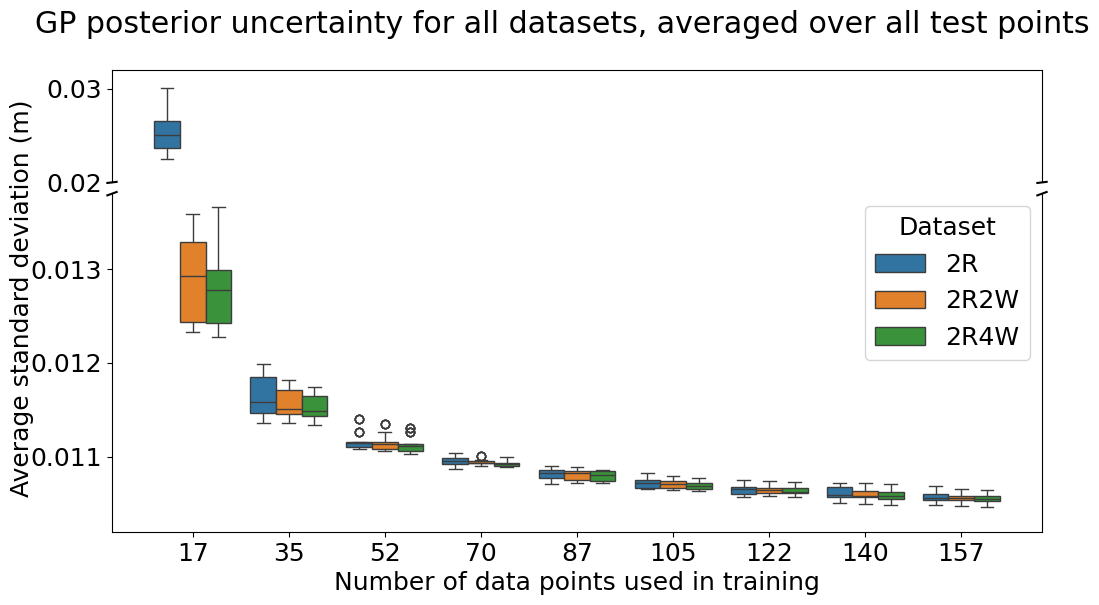}
\caption{The average posterior uncertainty for all the runs of the three datasets.}
\label{fig:UQ}
\end{figure}

Working with a GP has the benefit of being able to perform UQ. The posterior prediction for the $xyz$-coordinate consists of a mean and a variance. This variance can be interpreted as an uncertainty measurement. In this context, it serves as a quality assessment of the calibration. For every predicted 3D location, we have a posterior variance. In every run, we average over the standard deviations for the $x-$, $y-$, and $z-$coordinate for every predicted location. This gives us a mean standard deviation per 3D point. It is worth mentioning that this resulting uncertainty measurement can also be calculated by taking the RMSE of the three variances. However, in our context, the exact method used on how to combine uncertainties is not important, as the numbers are used in a relative way. The resulting posterior uncertainties of all runs are depicted in Figure \ref{fig:UQ}.

\begin{figure} 
    \centering
  \subfloat[Neural network predictions\label{1a}]{%
       \includegraphics[width=0.5\textwidth]{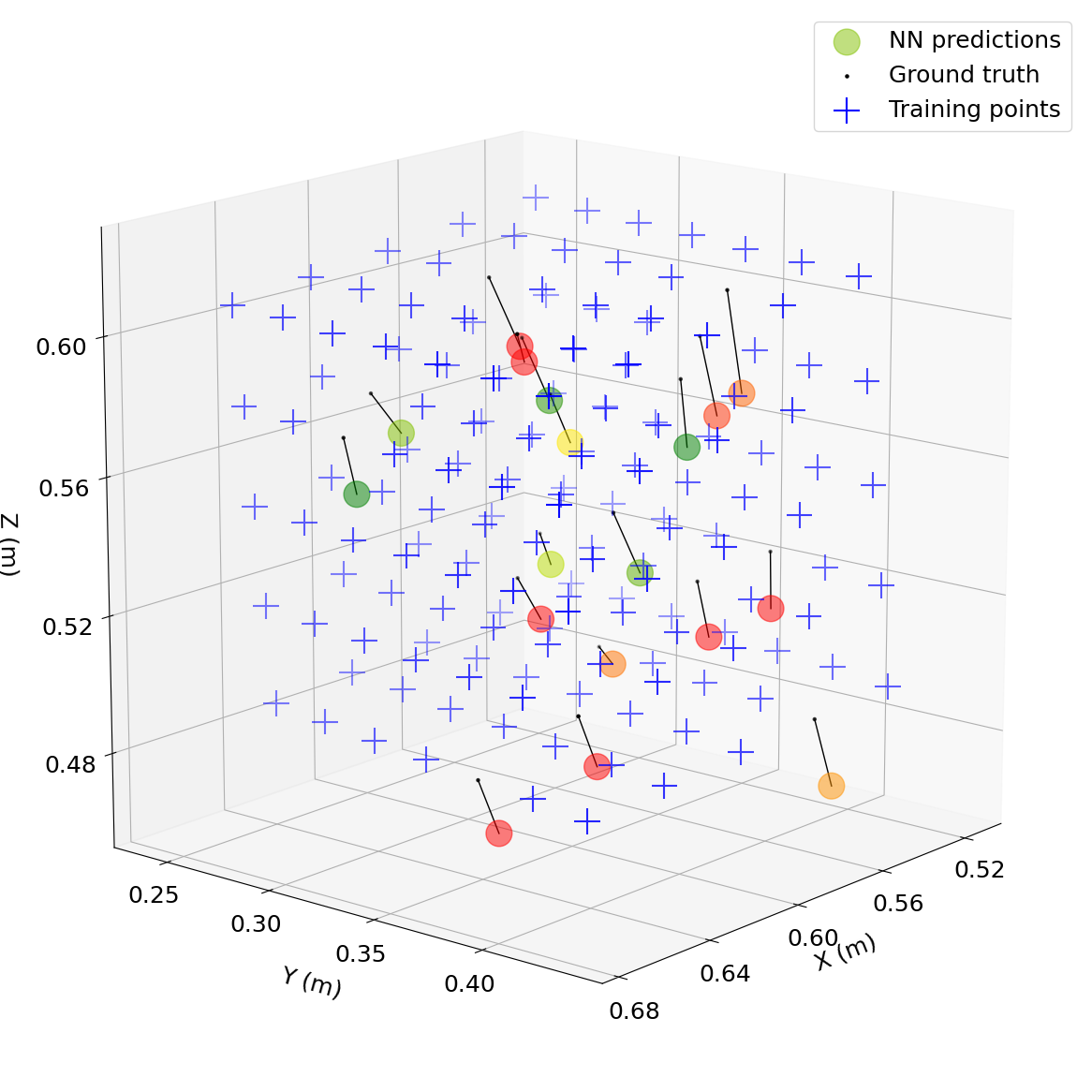}}
  \subfloat[Gaussian process predictions\label{1b}]{%
        \includegraphics[width=0.5\textwidth]{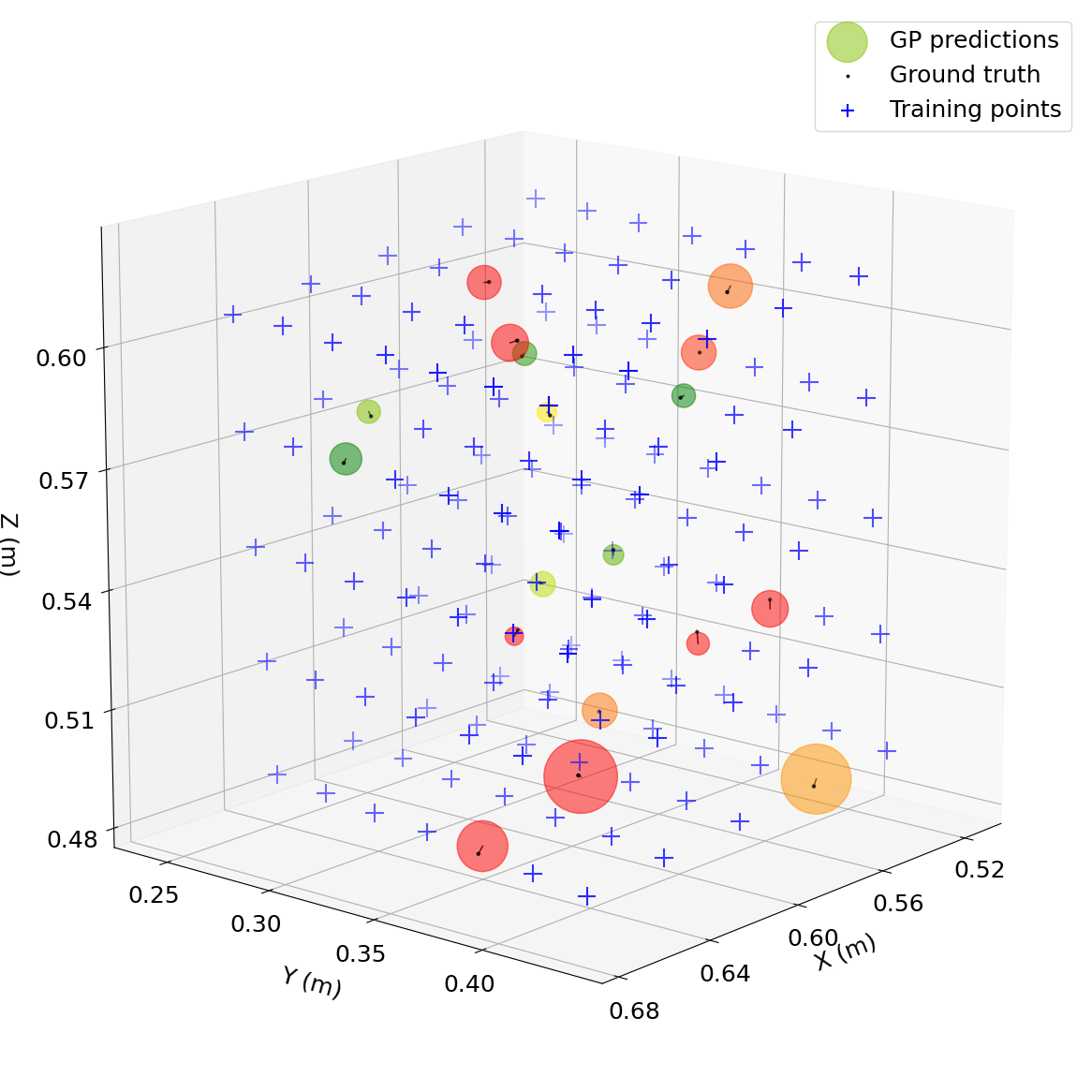}}
  \caption{Example of the predictions for a run of the 2R2W dataset with 157 3D locations by the neural network (a) and the Gaussian process (b). The light blue crosses are training points. The circles for the predictions are coloured between green (lowest RMSE) and red (highest RMSE). The Gaussian process predictions are scaled based upon their posterior variance. Smaller circles have less uncertainty than larger circles. The black dots are the ground truth. These plots are best visualised in 3D in an interactive window.} 
  \label{fig:UQExample} 
\end{figure}

Moreover, the UQ for the GP method can be visualised in a very informative way. In Figure \ref{fig:UQExample}, we show an example of a prediction for a run of the 2R2W dataset with 157 data points. This figure presents a comparative visualisation of 3D location predictions generated by the neural network (a) and the GP (b). The training data is visualised using light blue crosses. The true, correct locations that the models are attempting to predict are represented by small green circles (the ground truth). The model predictions themselves are shown as circles whose colour directly reflects their performance: predictions closer to the ground truth are green (lowest RMSE), while less accurate predictions are red (highest RMSE). A key feature of the GP plot (b) is that the size of the prediction circle is meaningful, as it is scaled by the model's posterior variance. This means that smaller circles indicate predictions where the GP is more certain (less uncertainty), and larger circles indicate predictions with greater uncertainty.

The comparative performance of the GP method consistently achieves lower RMSEs across all tested datasets compared to the neural network approach, as seen in Figure \ref{fig:RMSEs}. This is primarily due to the inherent data scarcity of the calibration task, a low-data regime where the GP, being a Bayesian model, effectively handles uncertainty and mitigates overfitting, a clear advantage over traditional neural networks that require more data for proper training.

The GP's performance is not uniformly better than the explicit camera calibration model. For the simpler 2R dataset, which provides only a four-dimensional input space based on the pixel coordinates of two cameras without a wide-angle lens, the GP's accuracy is worse because the input is insufficiently complex for the method to model the relationship effectively. In this specific scenario, the explicit model excels because it leverages the known mathematical structure of epipolar geometry, a powerful fixed prior. This remains true even when working with fewer images for the individual camera calibration and the stereo calibration. This is clearly visible in Figure \ref{fig:RMSEsa}. The green line plots the Root Mean Square Error (RMSE) for the explicit calibration using 50 images per camera and 10 images per stereo pair. The orange line corresponds to a calibration based on 25 images per camera and 6 per pair, while the red line represents the results for 10 images per camera and 4 images per pair.

Conversely, when input complexity increases, such as with the 2R2W (four cameras, 8D input, Figure \ref{fig:RMSEsb}) and the 2R4W dataset (six cameras, 12D input, Figure \ref{fig:RMSEsc}), the GP's flexibility proves beneficial. Here, the GP is able to outperform the explicit method, but only after surpassing a necessary threshold of data points, for instance, after 122 samples. This number is dependent on the chosen region of interest in 3D and the placement of the cameras. We also notice the significant deterioration of the explicit method for fewer boards (red lines). In this case, both data driven methods (GP and NN) outperform the explicit method when given enough data.

\subsection{Moving Ball with Active Learning}\label{ssec:MovingBallAL}

Of course, in the real world, ground truth data is not always available to construct images such as Figure \ref{fig:UQExample}. However we can still exploit the posterior uncertainty in AL, where new data points are gathered where the uncertainty is highest, thus optimizing the sampling procedure. This iterative procedure can be run until a satisfactory stopping criterion is met, for instance convergence in the total uncertainty or a limited amount of data points to be gathered.

\begin{figure}[!t]
    \centering
  \subfloat[AL dataset 2R\label{ALa}]{%
       \includegraphics[width=0.65\textwidth]{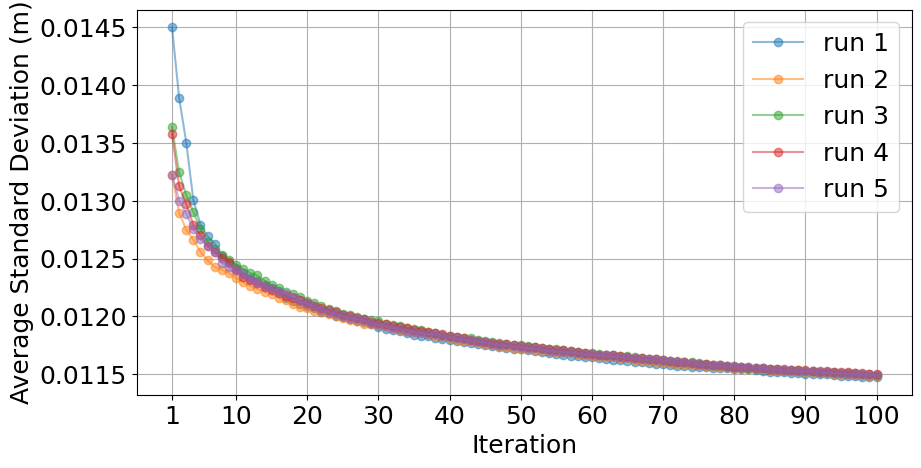}}
    \\
  \subfloat[AL dataset 2R2W\label{ALb}]{%
        \includegraphics[width=0.65\textwidth]{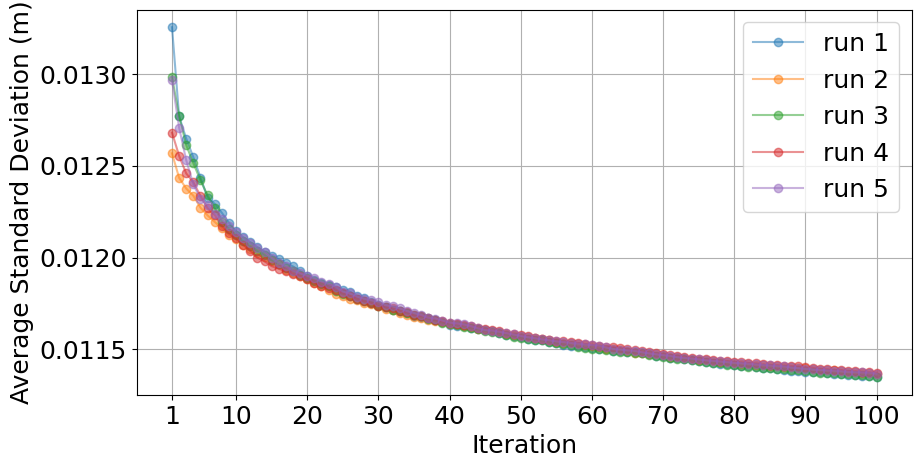}}
    \\
    \subfloat[AL dataset 2R4W\label{ALc}]{%
        \includegraphics[width=0.65\textwidth]{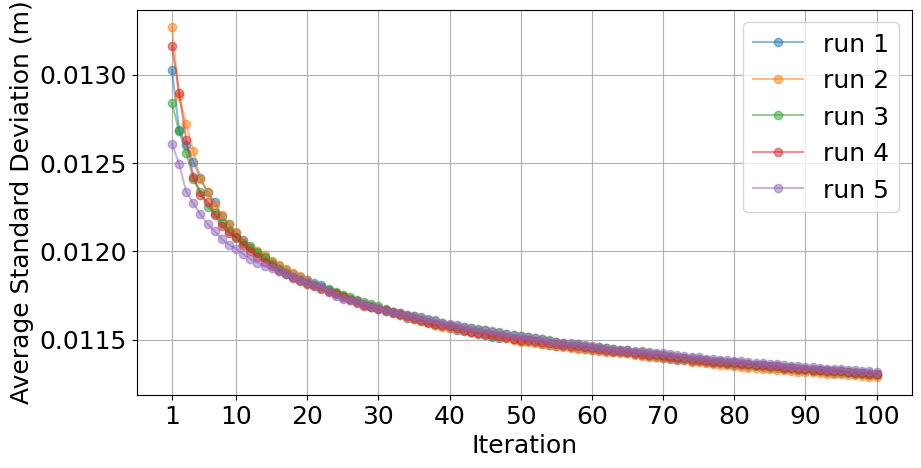}}
    \\
  \caption{The Active Learning iterations for every dataset. As can be expected from Figure \ref{fig:UQ}, the uncertainty quickly drops. Moreover, all curves converge almost immediately.}
  \label{fig:AL} 
\end{figure}

To demonstrate this, we start with a small subset of the $5\times5\times7$ grid of 175 3D locations and train the Gaussian process on those. Then, we make predictions for all remaining test points and pick as a new point, the one with the highest uncertainty, following the uncertainty sampling method. Next, we add this to the dataset and retrain the Gaussian process. We iteratively keep adding more points from the ground truth to the dataset. In practice, we do not have this ground truth, but pick points from a subset of the entire input domain, which are the $uv$-coordinates of the cameras in the multi-camera system to be calibrated. We start with 20\% of the 175 data points at random, and perform 100 iterations of Active Learning. We repeat this 5 times for every dataset to assess the dependants on the starting points that were chosen. The resulting plots are given in Figure \ref{fig:AL}.

Important to note, is that this new sample point is to be picked in pixel space, not in 3D space. In practice, the acquisition function is calculated for a dense grid of inputs, as it is cheap to evaluate. The location of the maximum of that acquisition function will serve as a new input point. When working with the UR10 robot, this means manually aligning the ball in the images in such a way, that the resulting $uv$-coordinates correspond to the coordinates determined by the location of the maximum of the acquisition function. However, this is relatively straightforward. The detection of the ball can be done in real time, even for six cameras in parallel.

Analysis of the UQ output in Figure \ref{fig:UQ} reveals several insights: Incorporating a higher number of cameras consistently yields lower posterior uncertainties, which is expected as more information is provided. Furthermore, uncertainty shows a rapid drop as the number of training data points increases, though using too few data points (e.g., 17 samples) prevents the GP from being trained correctly. Importantly, the resulting uncertainty curves in Figure \ref{fig:AL} demonstrate that the UQ is not dependent on the initial set of points chosen, as the model converges quickly, indicating a robust learning process.

\subsection{Moving Checkerboard}\label{ssec:MovingCheckerboard}

\begin{figure}[!t]
\centering
\includegraphics[width=0.65\textwidth]{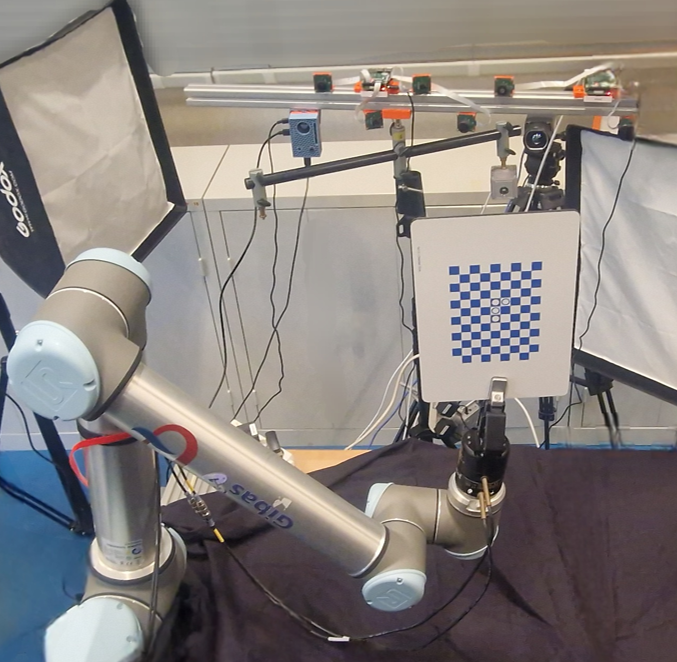}
\caption{A checkerboard mounted on a UR10 robot. This is translated by the robot perpendicular to itself to obtain a grid of known 3D coordinates of the corners.}
\label{fig:MovCB}
\end{figure}

\begin{figure}[!t]
\centering
\includegraphics[width=0.995\textwidth]{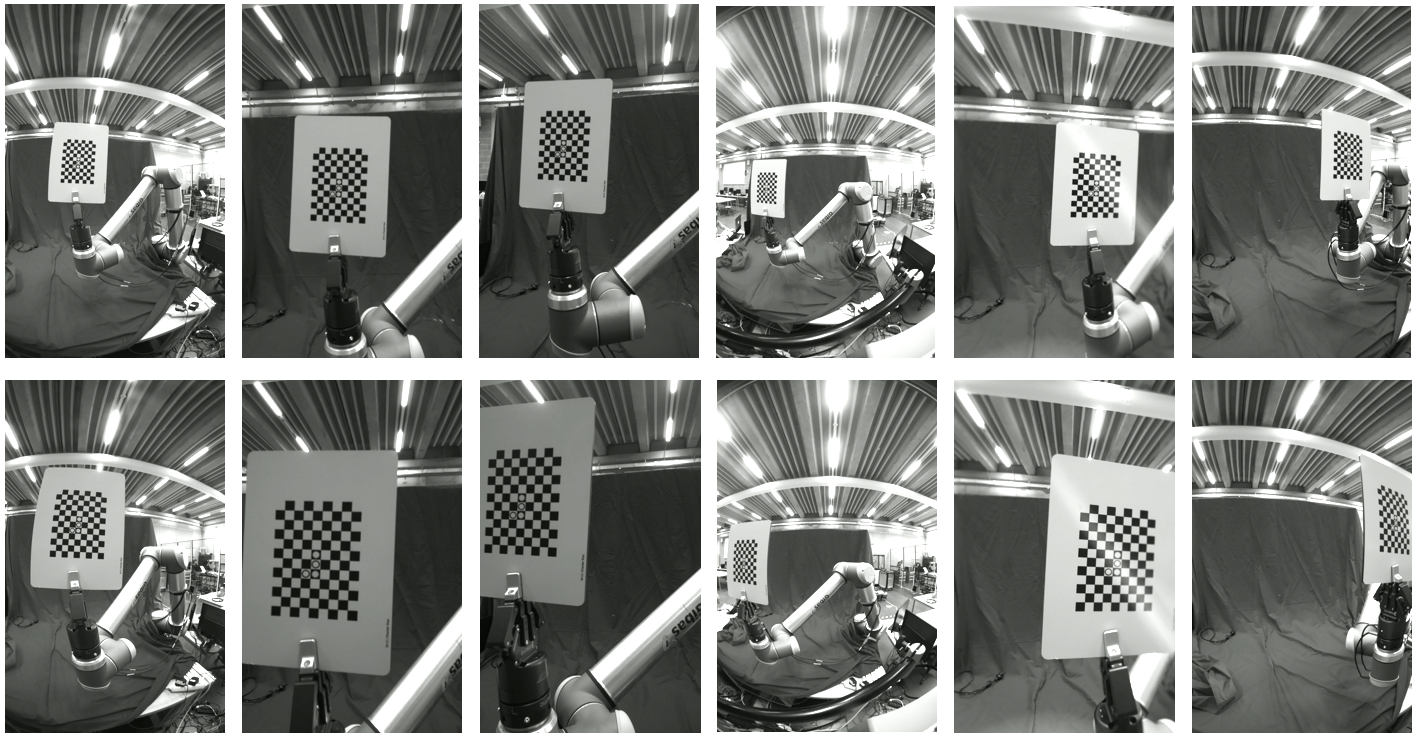}
\caption{Top row: the checkerboard in the starting position, as seen by the six cameras. Bottom row: the checkerboard translated to the last (closest) position.}
\label{fig:MCBPos}
\end{figure}

In a second experiment, we mount a checkerboard to the robot arm, as can be seen in Figure \ref{fig:MovCB}. We move the arm in such a way that the checkerboard is translated perpendicular to itself. We are interested in the ability of our method to predict 3D locations within the subregion traced out by these checkerboards. We want a dense positioning of data points in that region and are not interested in the regions outside that volume. The overall purpose is to construct a 3D measuring device that is calibrated with as few images as possible for a specific region. The main motivation for this experiment, is to see if we can drastically reduce the calibration of a multi-camera 3D scanner. This a tedious and time-consuming process.

We assign a 2D integer $xy$-coordinate to every corner of the checkerboard. The robot tracks the translation distance, which acts as the $z$-coordinate of the corners (we choose zero for the positions closest to the cameras, so increasing $z$-coordinate means further away). This yields the reference system for our setup. An overview of starting and ending images is given in Figure \ref{fig:MCBPos}. There are $8\times11$ corners on the checkerboard, which is placed in 20 positions. These form a dataset of $8\times11\times 20$ 3D locations for which the 2D $uv$-coordinates are found via corner detection using PyCBD~\citep{Hillen2023EnhancedProcesses}. We assess the capability of our method to reconstruct these 3D locations based on the images of the two regular cameras (2R), two regular cameras and 2 wide-angle lens cameras (2R2W) ans all six cameras (2R4W).

We train both the GPs and the neural net models on a subset of data points. We distinguish four scenarios for each of the three sets of cameras.  First, we keep only the closest and the furthest checkerboard in de training set and make predictions for the corners of all other checkerboards. This results in $8\times11\times 2=176$ training points. Next, we gradually include a checkerboard between the ones we already included until we have nine checkerboards in total, yielding $8\times11\times 9 = 792$ training points. 

\begin{figure*} [!t]
    \centering
    \subfloat[2R, 2 cb\label{MCBOverviewa}]{%
       \includegraphics[width=0.22\textwidth]{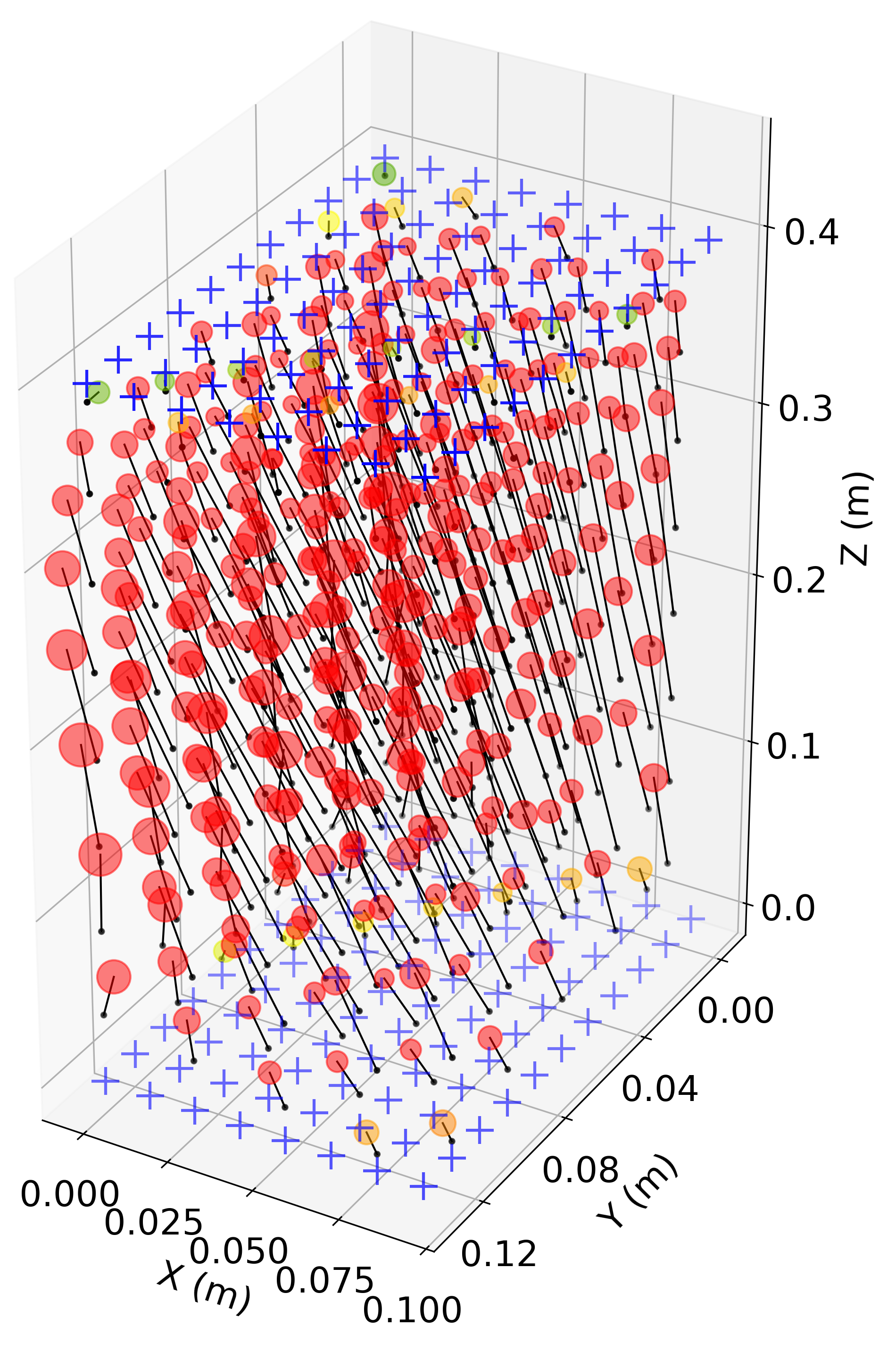}}
    \subfloat[2R, 3 cb\label{MCBOverviewd}]{%
       \includegraphics[width=0.22\textwidth]{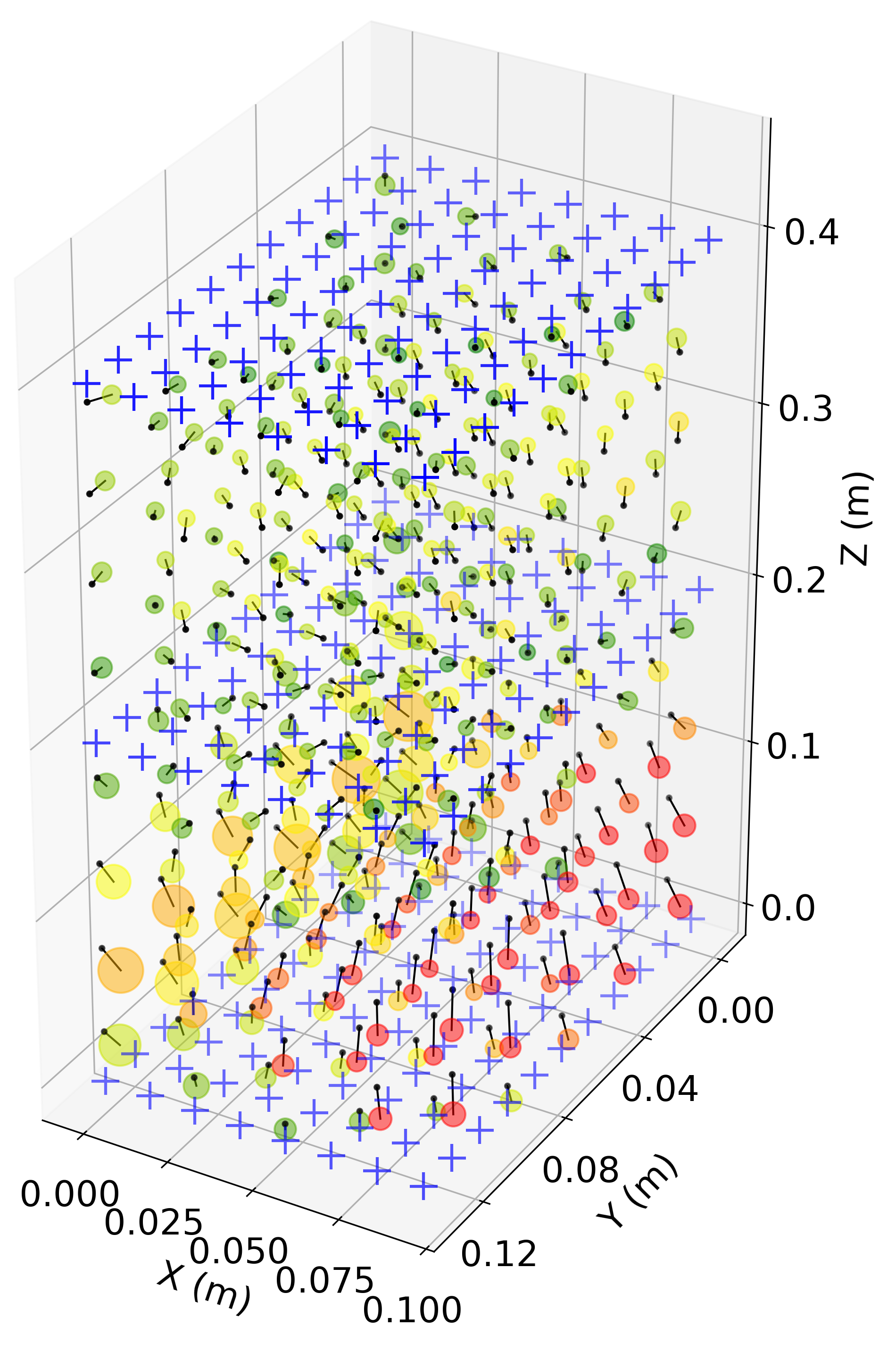}}
    \subfloat[2R, 5 cb\label{MCBOverviewg}]{%
       \includegraphics[width=0.22\textwidth]{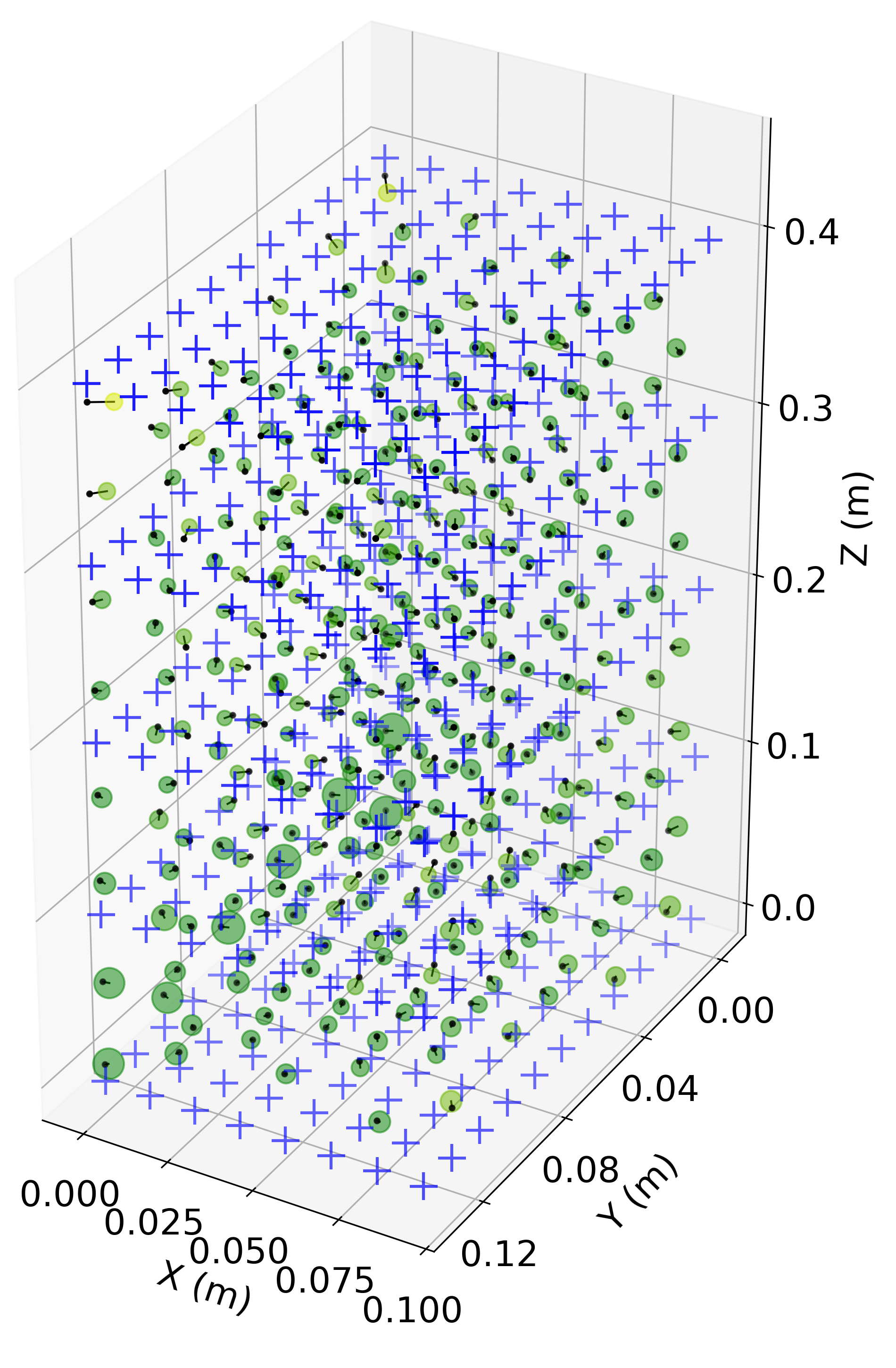}}
    \subfloat[2R, 9 cb\label{MCBOverviewj}]{%
       \includegraphics[width=0.22\textwidth]{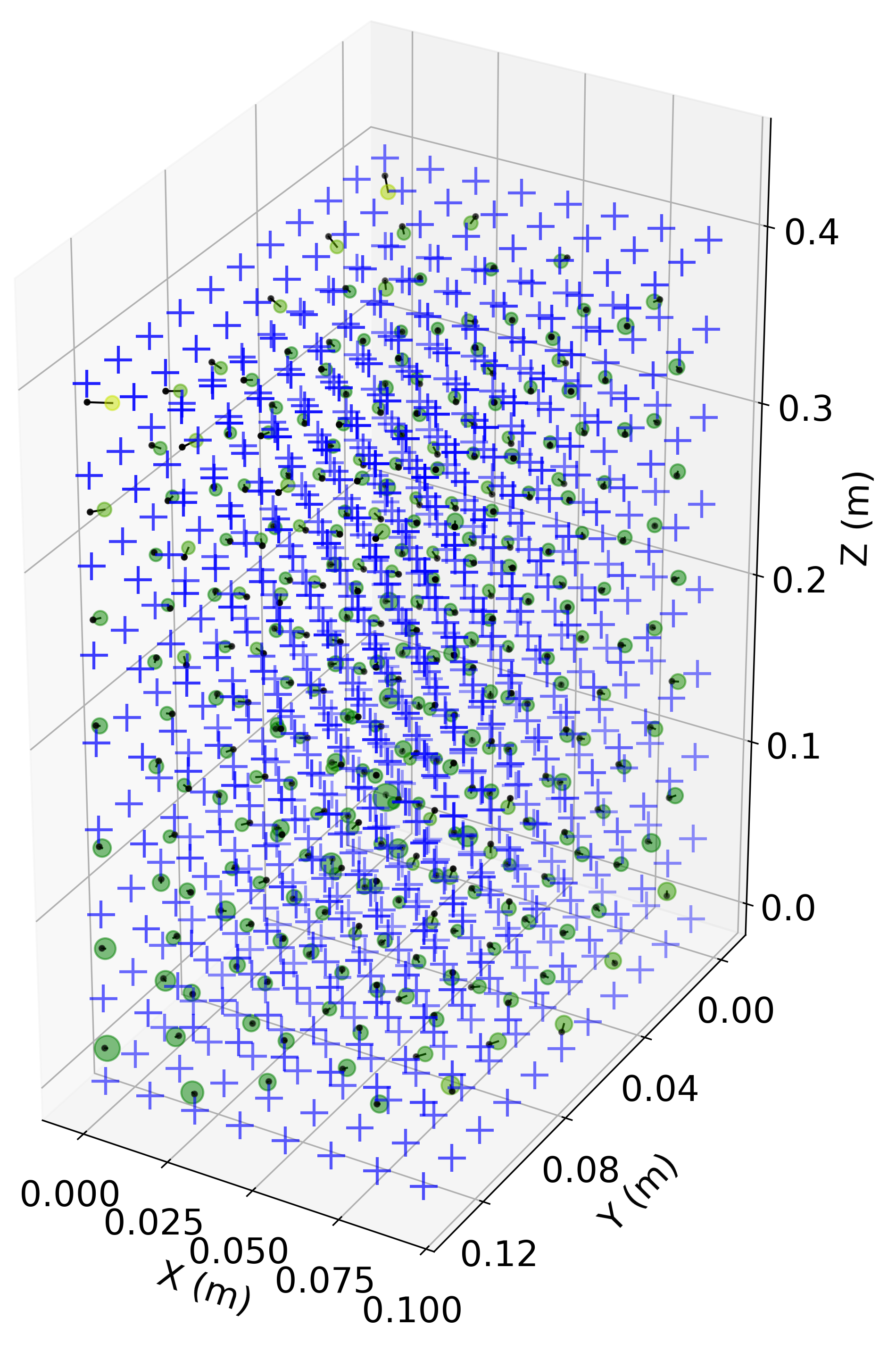}}
\\       
    \subfloat[2R2W, 2 cb\label{MCBOverviewb}]{%
       \includegraphics[width=0.22\textwidth]{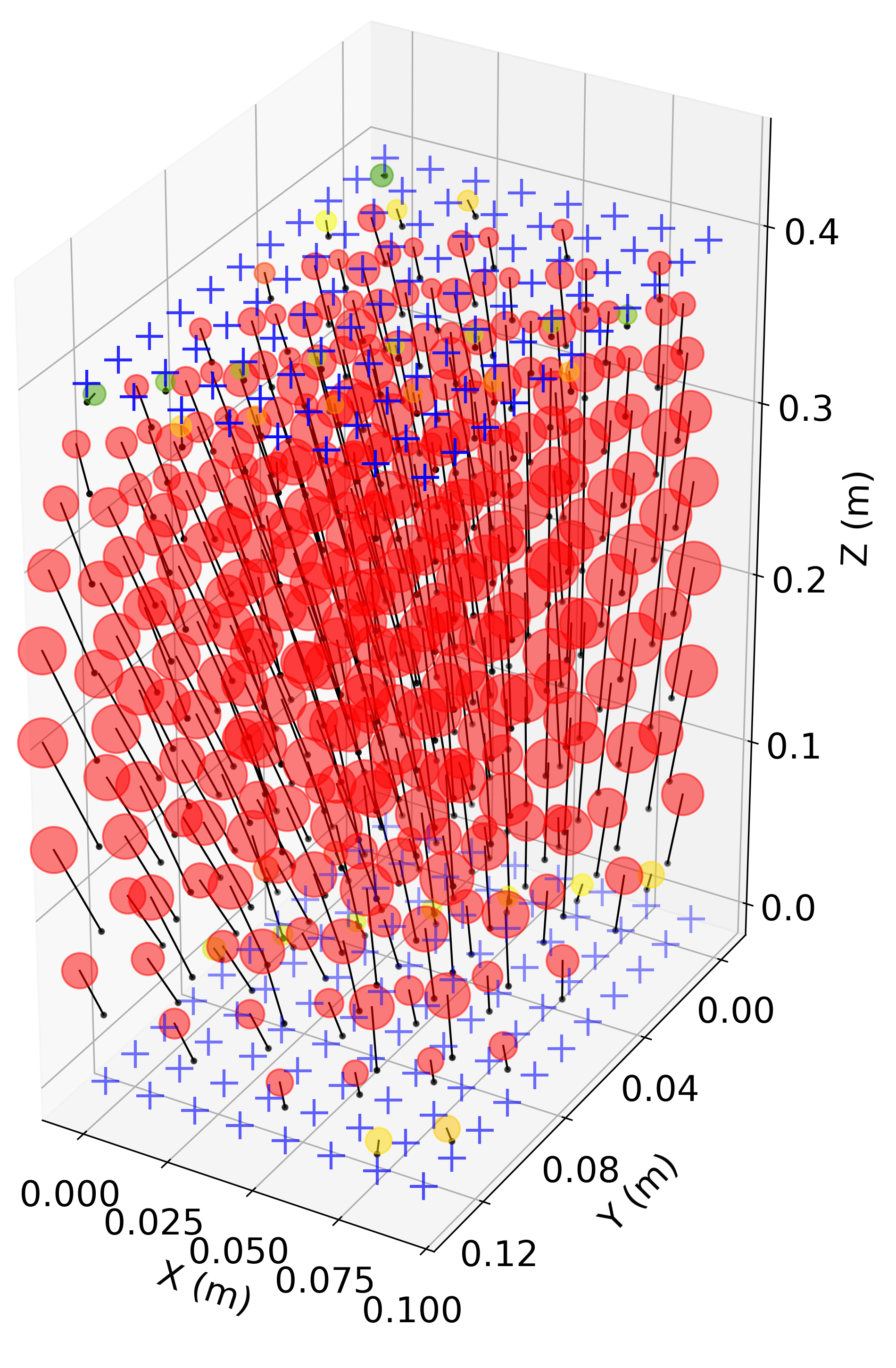}}
    \subfloat[2R2W, 3 cb\label{MCBOverviewe}]{%
       \includegraphics[width=0.22\textwidth]{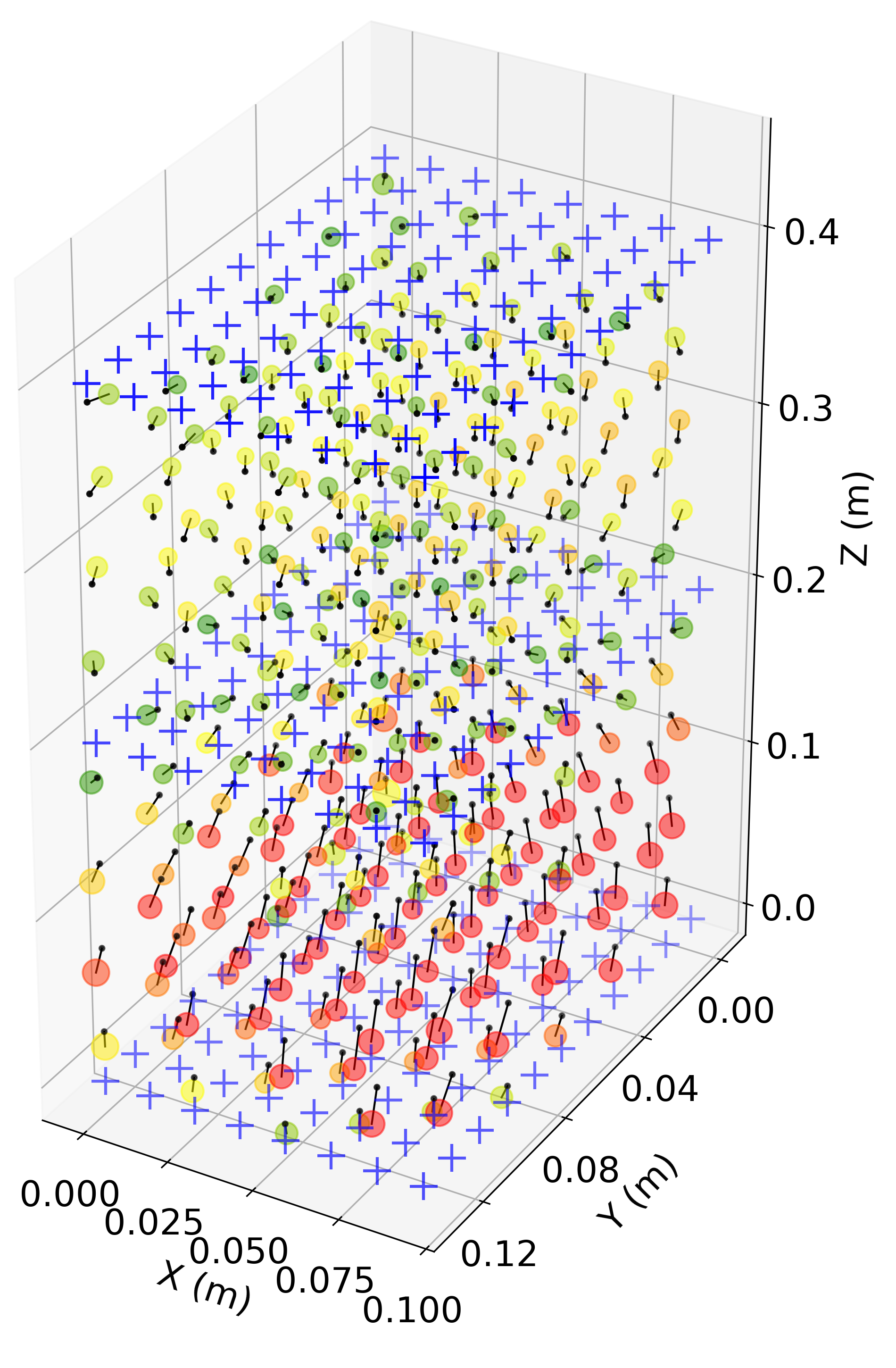}}
    \subfloat[2R2W, 5 cb\label{MCBOverviewh}]{%
       \includegraphics[width=0.22\textwidth]{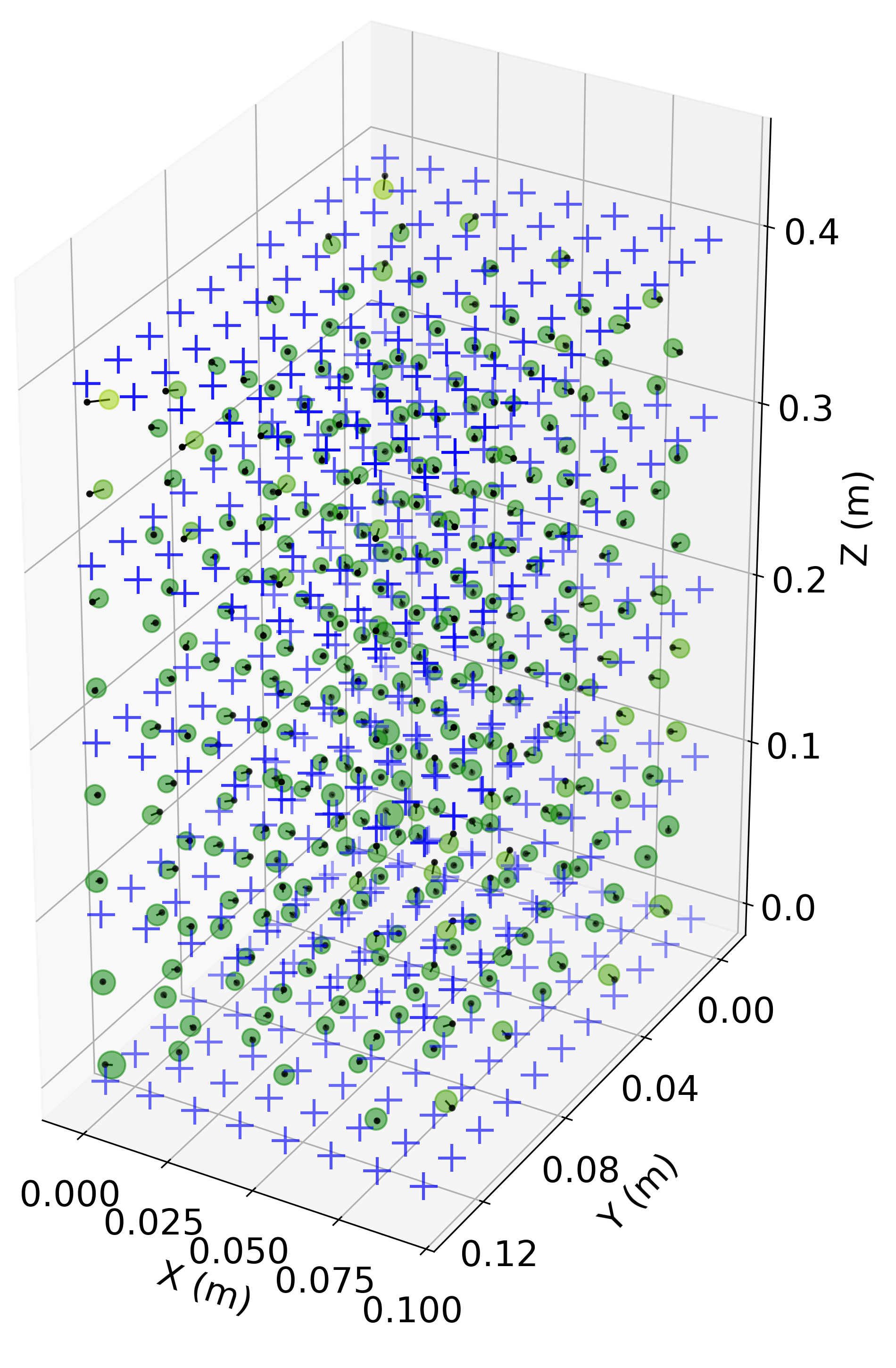}}
    \subfloat[2R2W, 9 cb\label{MCBOverviewk}]{%
       \includegraphics[width=0.22\textwidth]{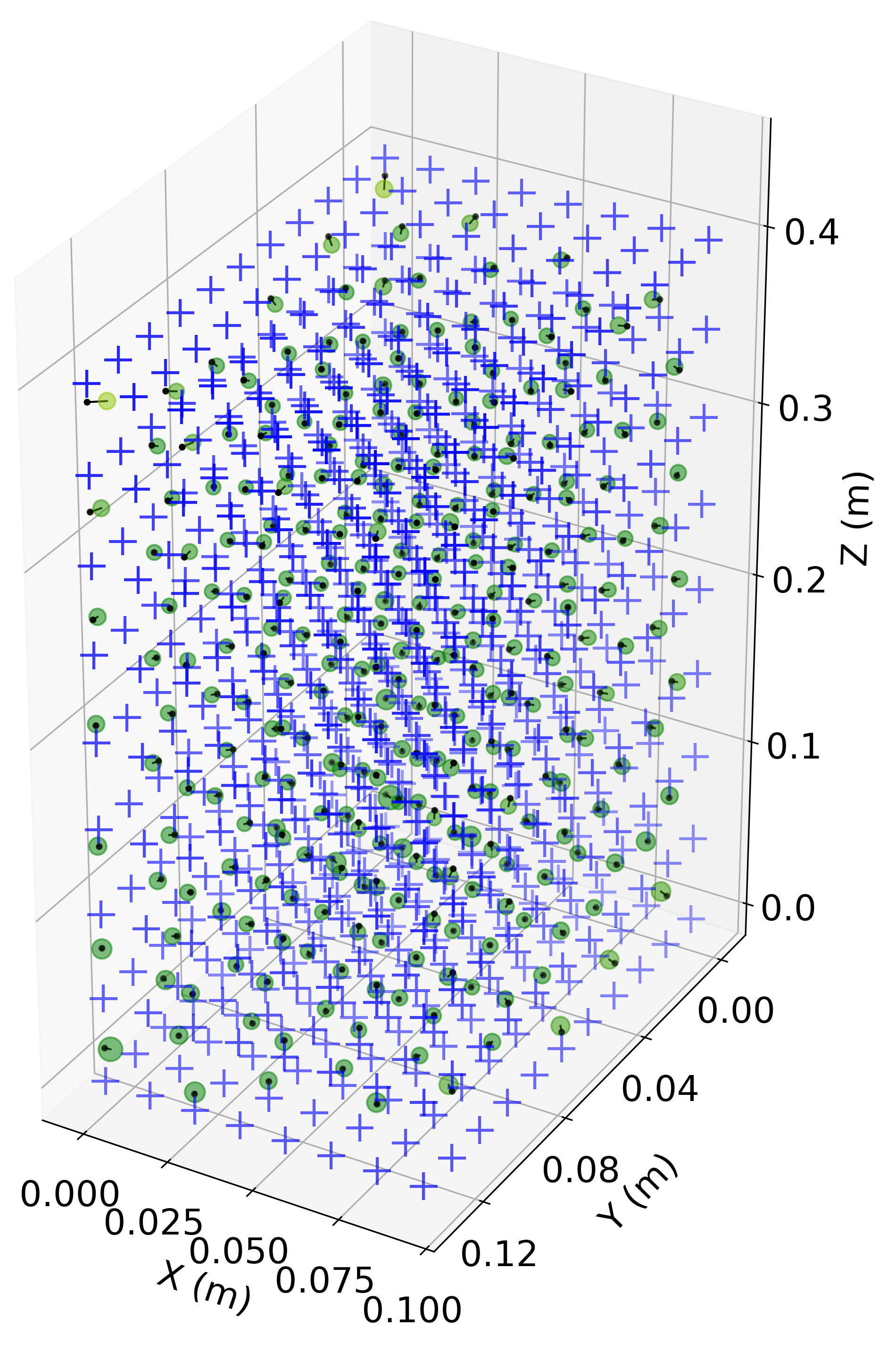}}
       \\
    \subfloat[2R4W, 2 cb\label{MCBOverviewc}]{%
       \includegraphics[width=0.22\textwidth]{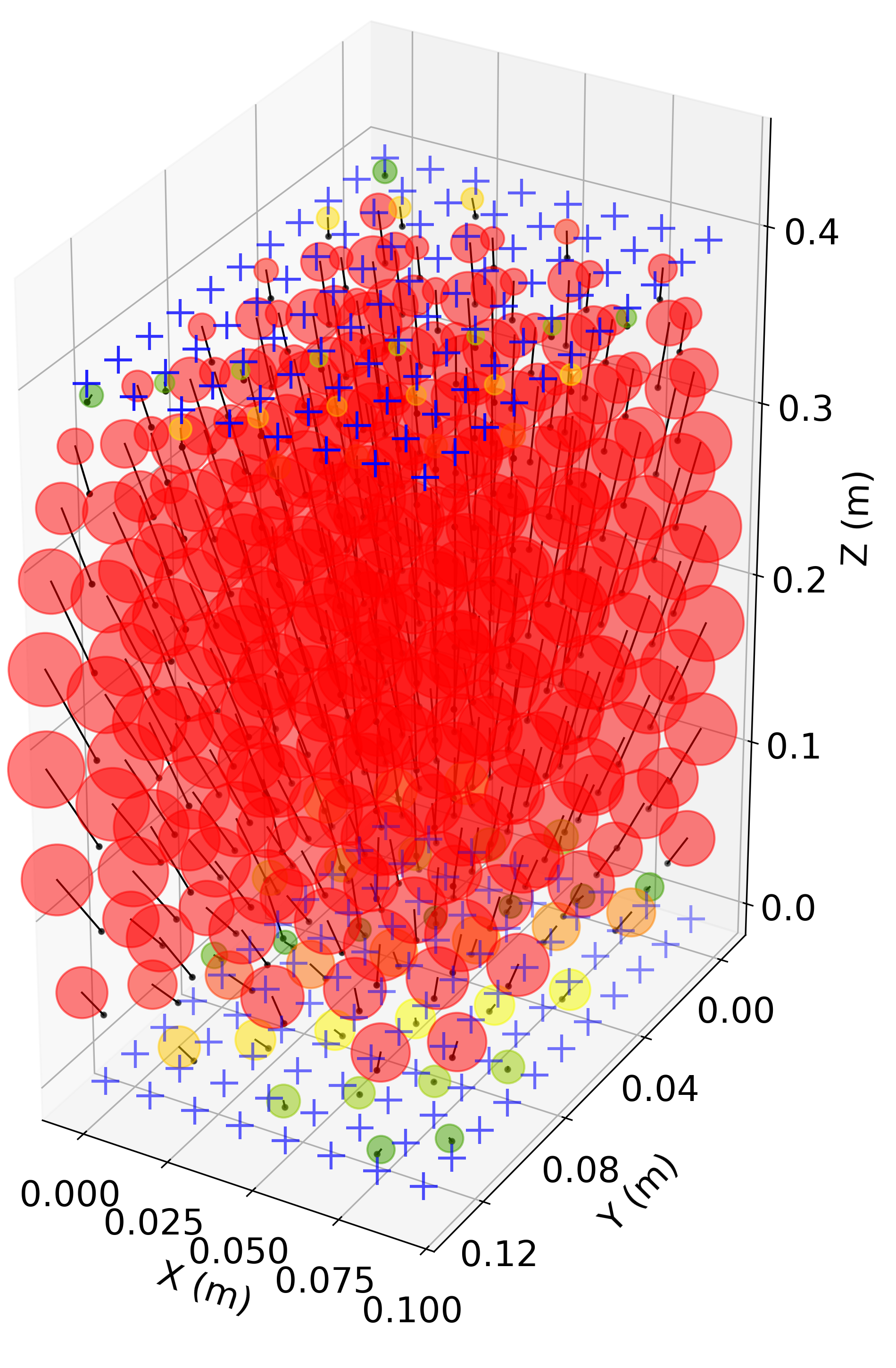}}    
    \subfloat[2R4W, 3 cb\label{MCBOverviewf}]{%
       \includegraphics[width=0.22\textwidth]{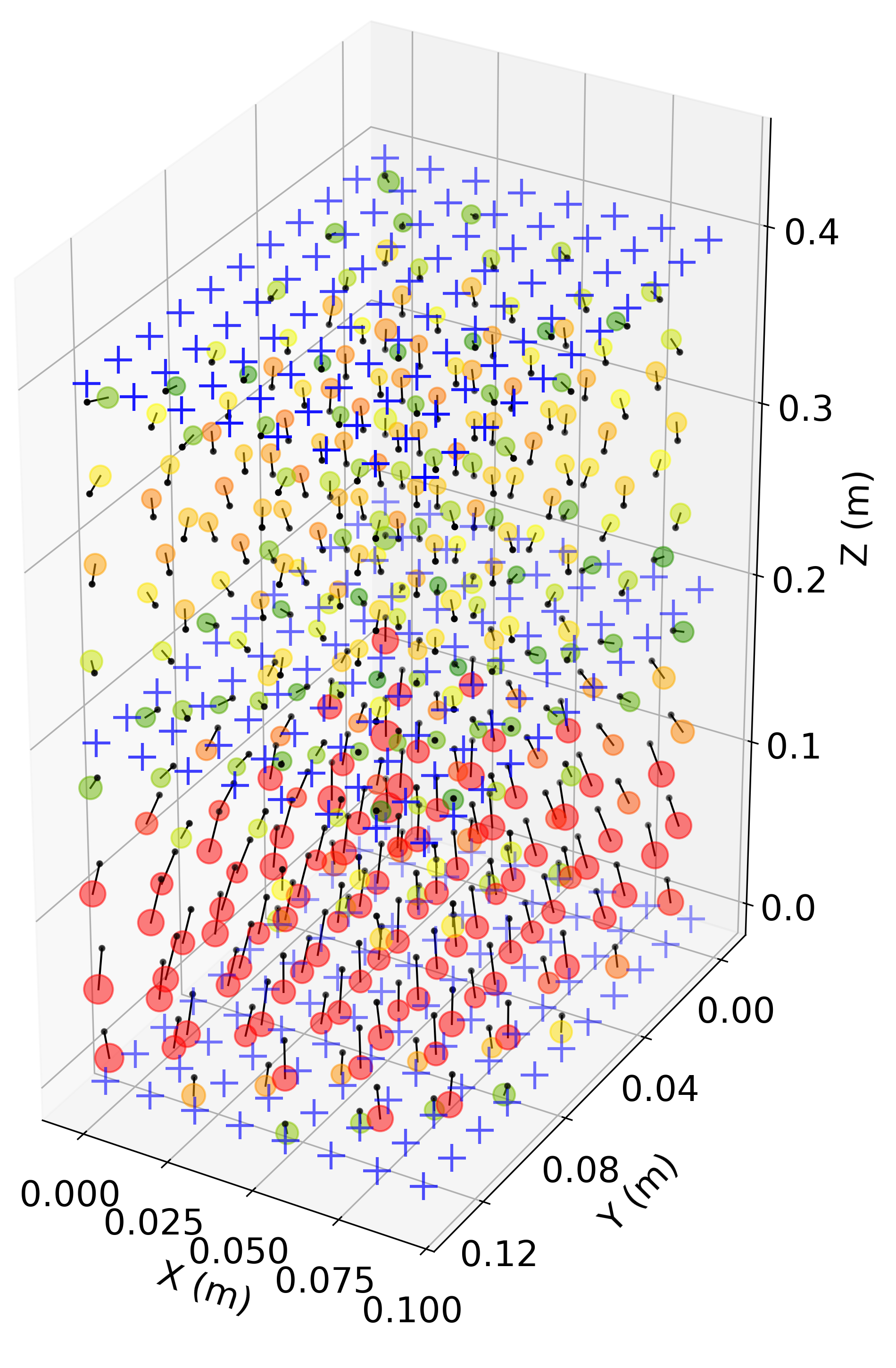}}    
    \subfloat[2R4W, 5 cb\label{MCBOverviewi}]{%
       \includegraphics[width=0.22\textwidth]{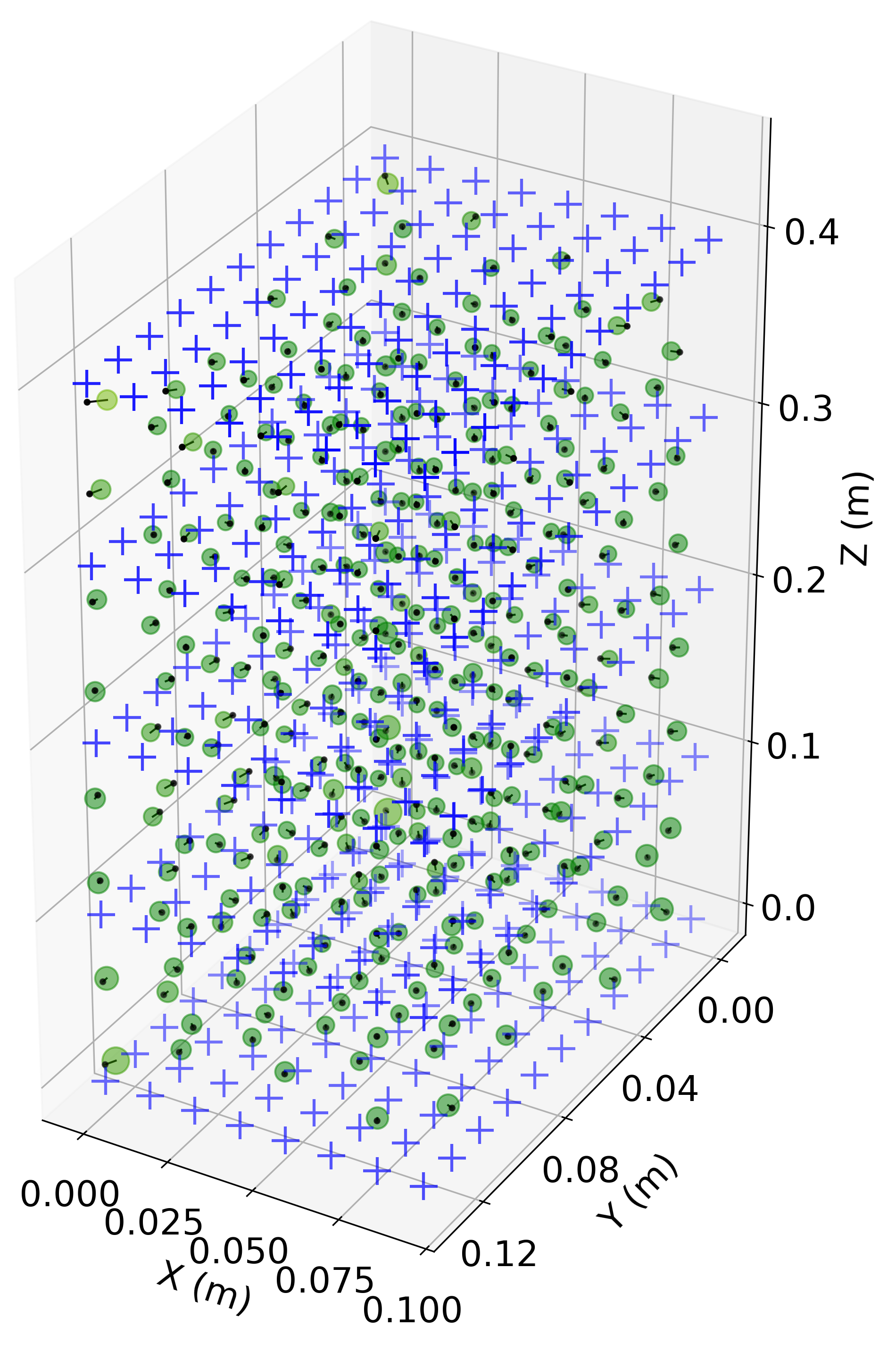}}    
    \subfloat[2R4W, 9 cb\label{MCBOverviewl}]{%
       \includegraphics[width=0.22\textwidth]{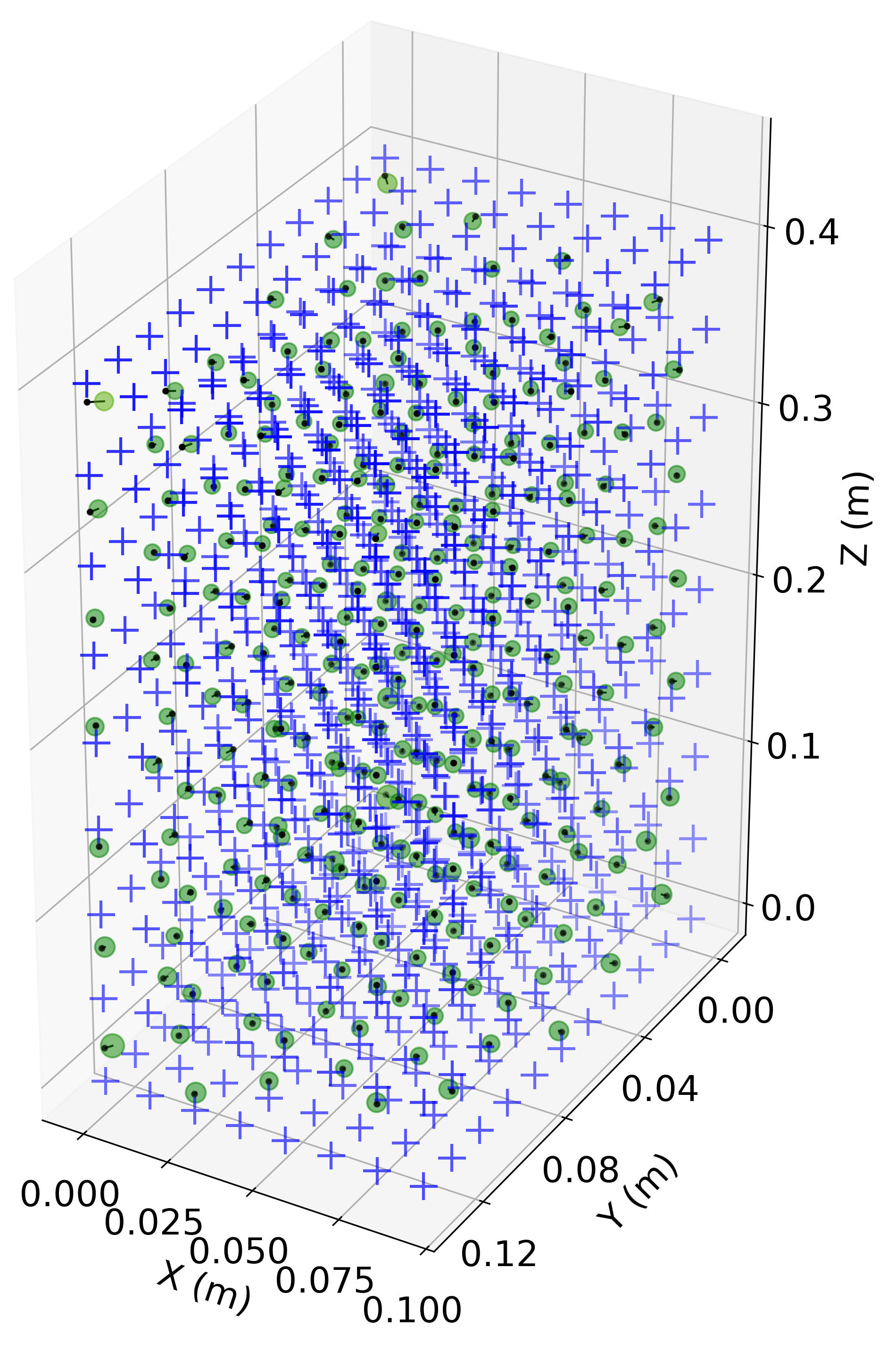}}
  \caption{Predictions of the GP models for the moving checkerboard experiment for the different datasets (top to bottom: 2R, 2R2W and 2R4W) and increasing number of checkerboards included in the training set (cb, left to right: 2, 3, 5, 9). The corners of the checkerboards are indicated with a blue cross. The RMSE is indicated by the colour (green is lower, red is larger). The size of the circle represents the uncertainty (larger is more uncertain). This plot is best viewed zoomed in.}
  \label{fig:MCBOverviewWide} 
\end{figure*}

An overview for the GP predictions is given in Figure \ref{fig:MCBOverviewWide}. We rescale the constructed integer coordinate system to real world coordinates by multiplying the $xy$-coordinates by the actual checkerboard square size of 13.29 mm. We multiply the $z-$coordinate by the step size of the robot arm, which is 10 mm. This means the coordinates in Figure \ref{fig:MCBOverviewWide} are real world units, with origin the first corner of the checkerboard in the first position.

From left to right, we include more checkerboards in the training set. From top to bottom, we compare the results for the different camera sets. The data points are depicted with blue crosses. The predictions are given by coloured circles. The colour represents the RMSE for that prediction with regards to ground truth. We use the same scale for all plots in order to make them comparable. A RMSE of zero metres corresponds to a green colour. A RMSE of 0.1 metres corresponds to a red colour. The size of the circles is determined by the posterior uncertainty, ranging from 7 to 30 for standard deviations from 0.01 metres to 0.1 metres respectively. These numbers are chosen for visualisation purposes. For clarity, we only plot one in every ten test points. We also plot a small black line between the ground truth and the predictions for those test points. We do not make a comparison against Zhangs' method. When all checkerboards are perfectly parallel, this will lead to an unreliable or impossible solution for the intrinsic camera parameters~\citep{Zhang2000ACalibration}.

A numerical analysis is given in Table \ref{tab:MCB}. We also provide a comparison against the neural network predictions. We worked with the same architecture as in Section \ref{ssec:MovingBallGrid}. Again, we can see that the GP method always outperforms the neural network model. Adding more checkerboard lowers both the RMSE and the posterior uncertainty. However, the last and also largest increase in number or checkerboards, from 5 to 9, yields the lowers benefit, indicating that there is little to be gained by adding more checkerboards. We also notice that more cameras results in lower RMSE.

\begin{table}[]
\centering
\caption{Results of the moving checkerboard experiment}
\label{tab:MCB}
\begin{tabular}{@{}ccccccc@{}}
\toprule
           & \multicolumn{2}{c}{RMSE GP} & \multicolumn{2}{c}{Avg Std GP} & \multicolumn{2}{c}{RMSE NN} \\ \midrule
 &
  \begin{tabular}[c]{@{}c@{}}Mean\\ (mm)\end{tabular} &
  \begin{tabular}[c]{@{}c@{}}Std\\ (mm)\end{tabular} &
  \begin{tabular}[c]{@{}c@{}}Mean\\ (mm)\end{tabular} &
  \begin{tabular}[c]{@{}c@{}}Std\\ (mm)\end{tabular} &
  \begin{tabular}[c]{@{}c@{}}Mean \\ (mm)\end{tabular} &
  \begin{tabular}[c]{@{}c@{}}Std\\ (mm)\end{tabular} \\
  \midrule
2R, 2 cb   & 29.30        & 12.52        & 10.89          & 0.46          & 39.14        & 25.05        \\
2R, 3 cb   & 4.61         & 2.40         & 10.52          & 0.48          & 10.92        & 4.74         \\
2R, 5 cb   & 1.36         & 0.60         & 10.34          & 0.20          & 13.56        & 6.17         \\
2R, 9 cb   & \textbf{1.05}         & 0.52         & 10.22          & 0.09          & 6.91         & 3.19         \\
\midrule
2R2W, 2 cb & 27.67        & 11.80        & 12.10          & 1.09          & 37.30        & 21.00        \\
2R2W, 3 cb & 5.88         & 3.12         & 10.48          & 0.15          & 10.53        & 8.08         \\
2R2W, 5 cb & 1.07         & 0.49         & 10.37          & 0.10          & 7.99         & 4.92         \\
2R2W, 9 cb & \textbf{0.83}         & 0.41         & 10.29          & 0.07          & 8.78         & 3.51         \\
\midrule
2R4W, 2 cb & 20.51        & 9.61         & 14.86          & 2.69          & 23.06        & 11.71        \\
2R4W, 3 cb & 6.79         & 3.54         & 10.52          & 0.19          & 12.23        & 9.71         \\
2R4W, 5 cb & 1.02         & 0.34         & 10.37          & 0.09          & 8.36         & 4.30         \\
2R4W, 9 cb & \textbf{0.65}         & 0.29         & 10.32          & 0.06          & 7.70         & 3.39         \\ \bottomrule
\end{tabular}%
\caption*{Std stands for standard deviation. Avg Std GP is the posterior uncertainty averaged over the three dimensions, for every test point. The number of checkerboards is given by cb.}
\end{table}

All datasets consistently exhibit greater uncertainty for regions located closer to the cameras. This is visually evident in Figures \ref{fig:MCBPos} (g) through (i), where points with lower z-coordinates are represented by larger circles. Correspondingly, there is an observable increase in RMSE (indicated by colours closer to red). This phenomenon is attributed to the fixed translation step of the checkerboard: when the checkerboard is closer to the camera, this fixed physical step results in a larger displacement in pixel space. Conversely, locations farther from the camera produce images that change little across the fixed translations, leading to a denser set of pixel input points in the dataset for those distant regions. This is an interesting result for people performing the calibration. Placing multiple parallel checkerboards equidistant, is suboptimal. There is more information to be gained by placing more checkerboards closer to the camera.

\section{Discussion}\label{sec:Discussion}

We omit a detailed analysis of the models' computational run-time because it is not the limiting factor in the overall calibration process. While both the explicit and implicit models execute within seconds due to the low-data regime, the biggest time bottleneck is the human-operated data acquisition (capturing the checkerboard images per camera and per stereo pair), which typically takes hours for larger multi-camera systems. Figure ~\ref{fig:RMSEs} illustrates that the number of images becomes particularly important in multi-camera settings. For the ordinary stereo pair, these three settings yield almost identical performance for explicit camera calibration, whereas for the configuration with 2 regular and 4 wide-angle cameras the jump from 10 to 25 images has a clear impact, while the additional increase from 25 to 50 images has only a minor effect. Since model execution is a mere fraction of the total time, a detailed computational comparison provides no practical utility. Our proposed method addresses this bottleneck by using AL to reduce the required data budget, a process that was automated in our setup by sending trigger pulses to controlling Arduinos.

In this preliminary study, we worked with cameras that map a 3D region of interest visible in all cameras. A compelling avenue of future work is addressing areas visible in only a subset of cameras. GPs lend themselves naturally to handle this missing data problem in low data regimes~\citep{Lalchand2022GeneralisedInference, DeBoiSurfAppr2023}.

One limitation of this study is that we did not assess how camera placement affects the results. Our experiments used only the rather horizontal setup depicted in Figure \ref{fig:6Cams}, although we evaluated three subsets of these cameras to demonstrate generalizability. Future work will investigate the placement of a much larger number of cameras in a more evenly spaced 2D grid and a 3D dome.

A key requirement of our implicit method is the need for known correspondences between image $uv$-coordinates and real-world $xyz$-coordinates. We relied on the UR10 robot to provide these, but acknowledge that this solution is not always feasible. Alternative non-robotic approaches, such as using a fixed mechanical construction (e.g., a large cuboid with parallel slots for a checkerboard), could be employed.

The explicit method produces 3D reconstructions in a camera-centric reference frame, requiring ICP registration to the robot ground truth for RMSE computation. This post-hoc rigid alignment can absorb global translation and rotation discrepancies, potentially reducing measured error and masking certain parametrization biases. However, this is a shortcoming of the assessment and not the method itself. The resulting calibration is not altered by this.

In our experiments, the robot ground truth (UR10) provides high repeatability (±0.1 mm) for relative positioning within its kinematic chain, but absolute accuracy may be limited by uncompensated errors. Thus, our RMSE primarily reflects relative reconstruction quality rather than absolute metrology-grade positioning.

A natural extension of this work would be to replace the three independent GPs for the \(x\), \(y\) and \(z\) coordinates with a multi-output (multi-task) Gaussian process that can model correlations between these outputs. While such a co-regionalised GP might improve data efficiency in very sparse regimes by sharing information across coordinates, it would also introduce a larger covariance structure, more hyperparameters and higher computational cost, which could complicate optimisation and limit scalability without guaranteeing a substantial gain in accuracy.

\section{Conclusion}\label{sec:Conclusion}

The core conclusion of this investigation is that the Gaussian process regression model is a highly effective solution for implicit multi-camera calibration, particularly when limited data is the dominant practical constraint. We demonstrated this in three experiments: a grid of known 3D locations, the Active Learning setting and the translated checkerboards.

The GP consistently outperforms the neural network because its inherent Bayesian structure is much better suited to handling low-data regimes, where standard neural networks fail to train reliably. Crucially, the GP provides a reliable measure of uncertainty quantification, which is essential for enabling AL strategies. This is the key to reducing the time a human operator spends physically acquiring calibration images.

An interesting result is that the predicted 3D uncertainty is higher closer to the cameras. This is because the data points are more sparse in the image ($uv$-coordinate) space in that region, even though the coverage may be dense in 3D world space. Moving the checkerboard a little bit when close to the cameras results in a relatively large change in pixel coordinates of the detected checkerboards.

However, the GP is not always the best choice. In the simplest case of basic stereo-camera calibration, where the input is a minimal four-dimensional space ($2\times$ $uv$-coordinates of two cameras), the GP cannot always infer the underlying relationship as well as the explicit method, which benefits immensely from the fixed mathematical prior of epipolar geometry. When complexity increases to more than two cameras, the GP's ability to successfully leverage that richer data structure allows it to surpass the explicit method, demonstrating its superior scalability for non-standard, high-dimensional systems.

Analysis of the posterior uncertainty confirmed the system's robustness: adding more cameras consistently yields lower uncertainties, and uncertainty drops rapidly as the number of data points increases. Furthermore, the final convergence point of the uncertainty curves is independent of the initial data set, confirming the stability of the GP algorithm.

Finally, while the computational time for all models is negligible (seconds), the true bottleneck is the hours required for the human operator to physically acquire images. The GP's real advantage lies in its ability to deliver an accurate, uncertainty-aware solution that enables intelligent data collection, directly addressing this major time-consuming aspect of the physical calibration process.

\backmatter

\bmhead{Data Availability Statement}
The data supporting the findings of this study are available upon reasonable request from the corresponding author.

\bmhead{Acknowledgements}
\noindent Ivan De Boi is funded by the FWO postdoc fellowship 1217125N and the University of Antwerp BOF-KP ID 53351.\\

\bibliography{references}

\end{document}